\documentclass[twoside,leqno,twocolumn]{article}

\usepackage[letterpaper]{geometry}

\usepackage{ltexpprt}
\usepackage{hyperref}

\usepackage[utf8]{inputenc} % allow utf-8 input
\usepackage[T1]{fontenc}    % use 8-bit T1 fonts
\usepackage{booktabs}       % professional-quality tables
\usepackage{amsfonts}       % blackboard math symbols
\usepackage{nicefrac}       % compact symbols for 1/2, etc.
\usepackage{microtype}      % microtypography
\usepackage{xcolor}        
\usepackage{tikz}
\usepackage{pgfplots}
\usepackage{xcolor}
\usepackage{amssymb}
\usepackage{wrapfig}
\usepackage{lipsum}
\usepackage{bm}

\usepackage{caption}
\usepackage{subcaption}

\usepackage{filecontents}
\usepackage{algorithm}
\usepackage{algorithmic}
\usepackage{algorithm}
\usepackage{wrapfig}
\usepackage{lipsum}
\usepackage{amsmath}
\usepackage{graphicx}
\usepackage{url}
\usepackage{amsfonts}
\usepackage{color, soul}
\usepackage{mathrsfs}
\usepackage{cleveref}
\usepackage{multirow}
\usepackage{float}
\usepackage{wrapfig}
\usepackage{bm}
\usepackage{bbm}
\usepackage{algorithm}
\usepackage{algorithmic}
\usepackage{colortbl}
\usepackage{enumitem}
\newtheorem{myDef1}{Problem}

\usepackage{multirow}

\begin{document}

\newcommand\relatedversion{}
\renewcommand\relatedversion{\thanks{The full version of the paper can be accessed at \protect\url{https://arxiv.org/abs/1902.09310}}} % Replace URL with link to full paper or comment out this line

%\setcounter{chapter}{2} % If you are doing your chapter as chapter one,
%\setcounter{section}{3} % comment these two lines out.

% debiasing knowledge distillation frameworks for GNNs 

\title{\Large RELIANT: Fair Knowledge Distillation for Graph Neural Networks}  % \relatedversion
\author{Yushun Dong\thanks{University of Virginia, Email: \{yd6eb, epb6gw, jundong\}@virginia.edu}
\and Binchi Zhang\footnotemark[1]
\and Yiling Yuan\thanks{Beijing University of Posts and Telecommunications, Email: yuanyiling@bupt.edu.cn}
\and Na Zou\thanks{Texas A\&M University, Email: nzou1@tamu.edu}
\and Qi Wang\thanks{Northeastern University, Email: q.wang@northeastern.edu}
\and Jundong Li\footnotemark[1]}

\date{}

\maketitle

% Copyright Statement
% When submitting your final paper to a SIAM proceedings, it is requested that you include
% the appropriate copyright in the footer of the paper.  The copyright added should be
% consistent with the copyright selected on the copyright form submitted with the paper.
% Please note that "20XX" should be changed to the year of the meeting.

% Default Copyright Statement
\fancyfoot[R]{\scriptsize{Copyright \textcopyright\ 2023 by SIAM\\
Unauthorized reproduction of this article is prohibited}}

% Depending on which copyright you agree to when you sign the copyright form, the copyright
% can be changed to one of the following after commenting out the default copyright statement
% above.

%\fancyfoot[R]{\scriptsize{Copyright \textcopyright\ 20XX\\
%Copyright for this paper is retained by authors}}

%\fancyfoot[R]{\scriptsize{Copyright \textcopyright\ 20XX\\
%Copyright retained by principal author's organization}}

%\pagenumbering{arabic}
%\setcounter{page}{1}%Leave this line commented out.

\begin{abstract} % \small\baselineskip=9pt 

Graph Neural Networks (GNNs) have shown satisfying performance on various graph learning tasks. To achieve better fitting capability, most GNNs are with a large number of parameters, which makes these GNNs computationally expensive. Therefore, it is difficult to deploy them onto edge devices with scarce computational resources, e.g., mobile phones and wearable smart devices.
Knowledge Distillation (KD) is a common solution to compress GNNs, where a light-weighted model (i.e., the student model) is encouraged to mimic the behavior of a computationally expensive GNN (i.e., the teacher GNN model). Nevertheless, most existing GNN-based KD methods lack fairness consideration. As a consequence, the student model usually inherits and even exaggerates the bias from the teacher GNN. 
To handle such a problem, we take initial steps towards fair knowledge distillation for GNNs. Specifically, we first formulate a novel problem of fair knowledge distillation for GNN-based teacher-student frameworks. Then we propose a principled framework named RELIANT to mitigate the bias exhibited by the student model.
Notably, the design of RELIANT is decoupled from any specific teacher and student model structures, and thus can be easily adapted to various GNN-based KD frameworks.
We perform extensive experiments on multiple real-world datasets, which corroborates that RELIANT achieves less biased GNN knowledge distillation while maintaining high prediction utility. Open-source code can be found at https://github.com/yushundong/RELIANT.
\end{abstract}

\vspace{2mm}

\section{Introduction}

%
% In recent years, network data has become ubiquitous over a plethora of domains, including online social networking~\cite{jiang2016understanding,alkhamees2021user} and e-commerce~\cite{xia2021self,zhou2021temporal}. In these examples, network data serves as an efficient representation of the relational data, where a node may encode complementary information about its neighbors.
% %
% To capture such complementary information and gain deeper understanding from these network data, Graph Neural Networks (GNNs) resort to the well-known message-passing mechanism. Correspondingly, they have shown satisfying performance in a variety of real-world applications, e.g., crime prediction~\cite{jin2020addressing}, medical diagnosis~\cite{saha2021graphcovidnet}, and credit risk scoring~\cite{wang2021temporal}, to name a few.

%  crime prediction~\cite{jin2020addressing}

In recent years, Graph Neural Networks (GNNs) have shown satisfying performance in a plethora of real-world applications, e.g., medical diagnosis~\cite{saha2021graphcovidnet} and credit risk scoring~\cite{wang2021temporal}, to name a few.
In practice, the depth and the number of parameters of GNNs largely determine their expressive power~\cite{he2022compressing}, which directly influence their performances in various graph learning tasks~\cite{chen2020simple}. 
Typically, deeper GNN layers enable the model to capture information that is multiple hops away from any node~\cite{DBLP:conf/iclr/KipfW17}, while a larger number of learnable parameters enable GNN to fit more complex underlying data patterns~\cite{chen2020simple}.
However, in most cases, the inference efficiency of GNNs is inevitably degraded by the deep layers or the large number of parameters.
Such efficiency degradation naturally makes these GNNs inapplicable to be deployed on edge devices (e.g., mobile phones) with limited computational resources~\cite{he2022compressing,joshi2021representation}.

\begin{figure}[!t]
\centering
    % \vspace{-1mm}
        \begin{subfigure}[t]{0.24\textwidth}
        \small
        \includegraphics[width=0.98\textwidth]{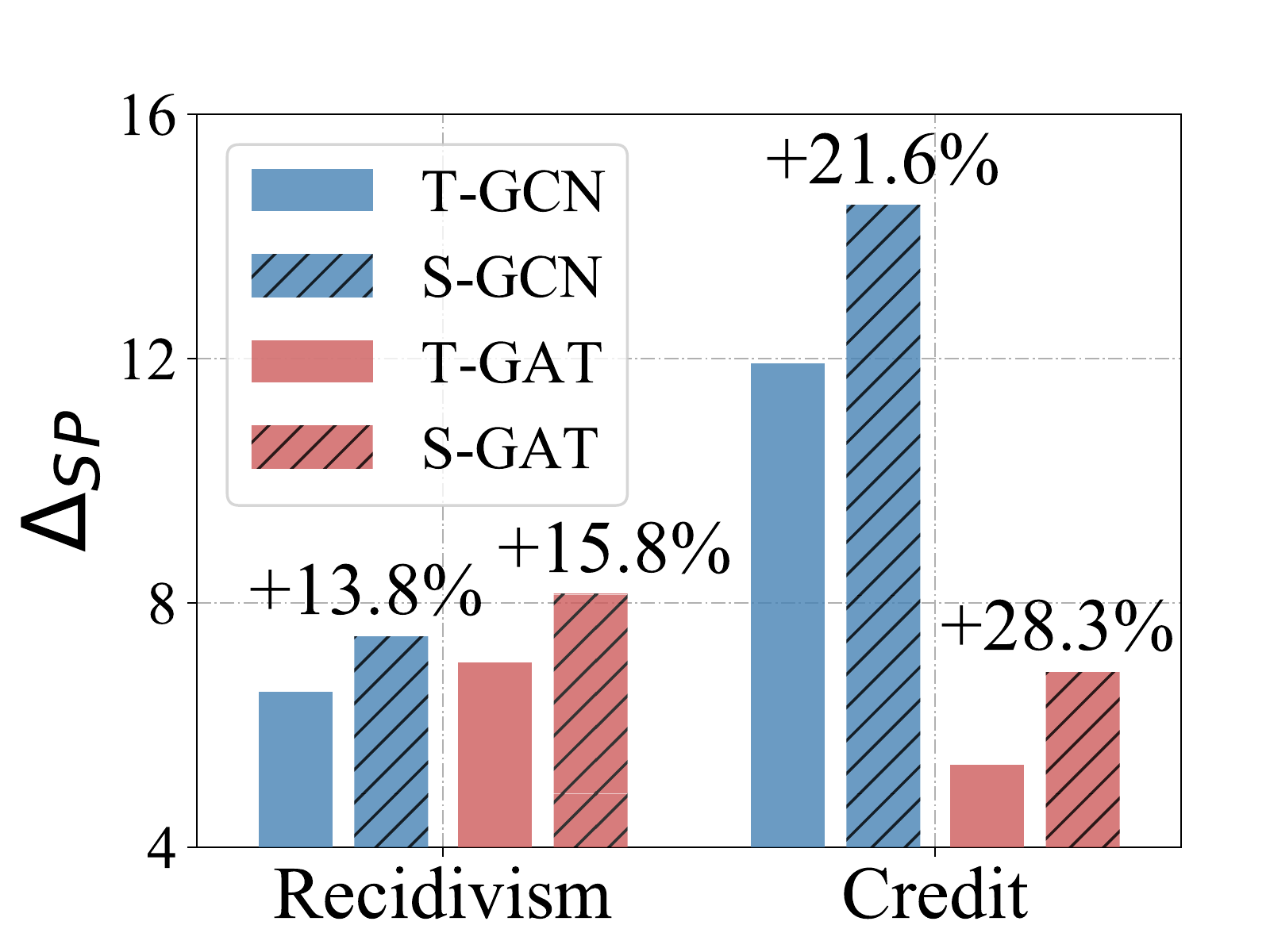}
        % \vspace{-1mm}
            \caption[Network2]%
            % {{\footnotesize $\Delta G_{\text{SP}}$ on Income}} 
            {{\footnotesize Bias under $\Delta_{\text{SP}}$ on CPF.}} 
            \label{pre_1}
        \end{subfigure}
                        \begin{subfigure}[t]{0.24\textwidth}
        \small
        \includegraphics[width=0.98\textwidth]{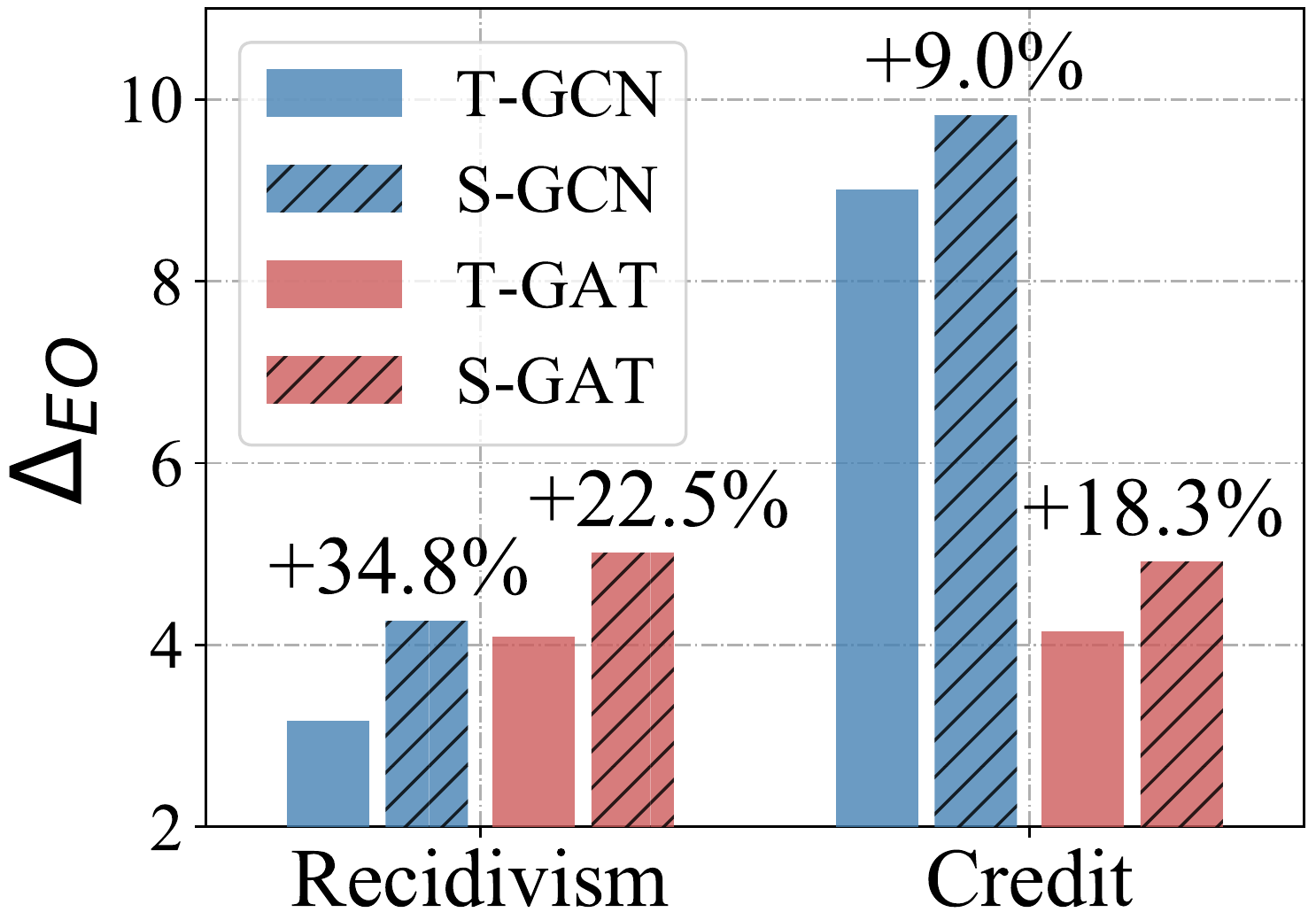}
        % \vspace{-1mm}
            \caption[Network2]%
            {{\footnotesize Bias under $\Delta_{\text{EO}}$ on AKD.}}    
            \label{pre_2}
        \end{subfigure} 
    \vspace{-2mm}
    \caption{A comparison of exhibited bias between teacher and student models based on two representative GNN knowledge distillation frameworks (CPF and GraphAKD). "T" and "S" represent the teacher and the student model, respectively. The names of GNN mark out the corresponding teacher models.}
    \vspace{-8mm}
    \label{preliminary_analysis}
\end{figure}

Due to the problem above, it is necessary to compress those computationally expensive GNNs for deployment on edge devices.
Knowledge Distillation (KD) is a common approach to compress GNNs but still maintains a similar level of prediction performance~\cite{yang2021extract,he2022compressing,joshi2021representation}. 
Here, the basic idea of KD is to let a light-weighted student model (as the compressed GNN) learn to mimic the behavior (e.g., output logits) of the teacher model (usually a computationally expensive GNN).
%
% The main idea is that the student network mimics the behavior of the teacher network to obtain a competitive or even a superior performance [14, 37], while the teacher network transfers soft tar- gets, hidden feature maps, or relations between pair of layers as distilled knowledge to the shallow student network.
%which has attracted growing attention in recent years.
%
% In recent years, the knowledge distillation based on GNNs has attracted growing attention due to its effectiveness in compressing GNNs for efficient inference.
%
However, most existing KD approaches do not have any fairness consideration over different demographic subgroups, and the optimized student model often preserves and even exaggerates the exhibited bias from the teacher GNN.
Consequently, when the compressed model is deployed in real-world application scenarios, there could exist discrimination toward specific populations.
Here we provide preliminary analysis based on two representative GNN knowledge distillation frameworks, namely CPF~\cite{yang2021extract} and GraphAKD~\cite{he2022compressing}.
Specifically, we measure the exhibited bias in the widely-studied node classification task on two real-world datasets. Here Recidivism is a network of defendants~\cite{jordan2015effect,agarwal2021towards}, while Credit is a network between bank clients~\cite{yeh2009comparisons,agarwal2021towards}.
We adopt two traditional metrics, i.e., $\Delta_{\text{SP}}$ (measuring the level of bias under Statistical Parity~\cite{dwork2012fairness}) and $\Delta_{\text{EO}}$ (measuring the level of bias under Equal Opportunity~\cite{hardt2016equality}), to measure the exhibited bias of GNN predictions.
We present a comparison of the exhibited bias between teacher and student models in Fig.~\ref{preliminary_analysis}. Empirical results show that student models tend to yield more biased results compared with the teacher GNN model, which could be attributed to the biased guidance from the teacher GNN during training.
It is worth noting that in most cases, directly retraining the teacher GNN for debiasing is infeasible, since retraining the teacher GNN with a large number of parameters is computationally expensive.
Hence, mitigating the bias for the student model is an urgent need.
%
% It is worth noting that in most human-centered applications, the decisions made could be life-changing for those involved individuals, especially under those high-stake scenarios, e.g., loan application and criminal justice.
%
% Hence, it is an urgent need to mitigate such bias exhibited by the student model during GNN knowledge distillation.

Despite the necessity of mitigating bias for the student model, existing exploration remains scarce.
In this paper, we aim to make an initial step towards developing a debiasing framework that can be easily adapted to various existing GNN-based KD methods.
However, this task is non-trivial mainly due to the following three challenges: 
% the biased teacher knowledge cannot be changed
% (1) \textbf{Gap towards Fair Knowledge:} For most KD frameworks designed for compressing GNNs, the parameters of teacher GNN is often frozen during the training of the student model due to the lack of supervision information for the teacher model. Therefore, if the teacher GNN exhibits any bias, such biased knowledge could be adopted as the supervision for the student model. Hence, learning a fair student model with biased supervision from the teacher GNN is our first challenge.
(1) \textbf{Gap towards Fair Knowledge:} For most KD frameworks designed for compressing GNNs, the teacher GNN model usually serves as the sole source of supervision signal for the training of the student model. Therefore, if the teacher GNN exhibits any bias, such biased knowledge tends to be inherited by the student model. Hence, learning a fair student model with biased supervision from the teacher GNN is our first challenge.
(2) \textbf{Gap towards End-to-End Learning:} A critical advantage of existing KD models is the end-to-end learning paradigm, which enables the distilled knowledge to be tailored to specific downstream tasks.
In such an end-to-end learning process, highly efficient gradient-based optimization techniques are widely adopted. However, widely-used fairness notions (e.g., Statistical Parity and Equal Opportunity) are defined on the predicted labels. Hence the corresponding bias metrics are naturally non-differentiable w.r.t. the student model parameters. 
Developing a debiasing framework suitable for gradient-based optimization techniques in the end-to-end learning paradigm is our second challenge.
(3) \textbf{Gap towards Generalization:} 
Various KD models have been proposed for compressing GNNs to satisfy different application scenarios. In fact, most KD models are developed based on certain designs of student models. Developing a framework that is student-agnostic and easily adapted to different KD models is our third challenge.

To tackle the above challenges, in this paper, we propose a novel framework named 
RELIANT (fai\underline{R} knowl\underline{E}dge disti\underline{L}lat\underline{I}on for gr\underline{A}ph \underline{N}eural ne\underline{T}works)
to mitigate the bias learned by the student model. Specifically, we first formulate a novel research problem of \textit{Fair Knowledge Distillation for GNN-based Teacher-Student Frameworks}.
To tackle the first challenge, we incorporate a learnable proxy of the exhibited bias for the student model. In this way, despite the knowledge (from the teacher GNN) being biased, the student model still makes less biased predictions under proper manipulations on the proxy.
To tackle the second challenge, we propose to approximate the bias level of the student model, where the approximation is differentiable (w.r.t. the student model parameters) manner. In this way, the highly efficient end-to-end learning paradigm is preserved, and the gradient-based optimization techniques are still applicable.
To tackle the third challenge, we design the proposed framework RELIANT in a student-agnostic manner. In other words, the debiasing for the student model does not rely on any specific design tailored for the student model structure. Therefore, RELIANT can be easily adapted to different GNN-based knowledge distillation approaches.
%
% Finally, we perform extensive experiments on multiple real-world network datasets to corroborate the effectiveness of RELIANT.
%
The main contributions of this paper are summarized as follows. %
\begin{itemize}[topsep=2pt]
    \item \textbf{Problem Formulation.} We formulate and make an initial investigation on a novel research problem of fair knowledge distillation for GNN-Based teacher-student frameworks.
    
    \item \textbf{Algorithmic Design.} We propose a principled framework named RELIANT that learns the proxy of bias for the student model during KD. RELIANT achieves student-agnostic debiasing via manipulating the proxy during inference.
    
    \item \textbf{Experimental Evaluation.} We conduct comprehensive experiments on multiple real-world datasets to verify the effectiveness of the proposed framework RELIANT in learning less biased student models.
\end{itemize}

\section{Problem Definition}

\noindent \textbf{Notations.}
We denote matrices, vectors, and scalars by bold uppercase letters (e.g., $\mathbf{X}$), bold lowercase letters (e.g., $\mathbf{x}$), and regular lowercase letters (e.g., $x$), respectively. For any matrix, e.g., $\mathbf{X}$, we use $\mathbf{X}_{i,j}$ to indicate the element at the $i$-th row and $j$-th column.

\begin{figure*}[!t]
    \centering
    \vspace{-4mm}
    \includegraphics[width=0.9\textwidth]{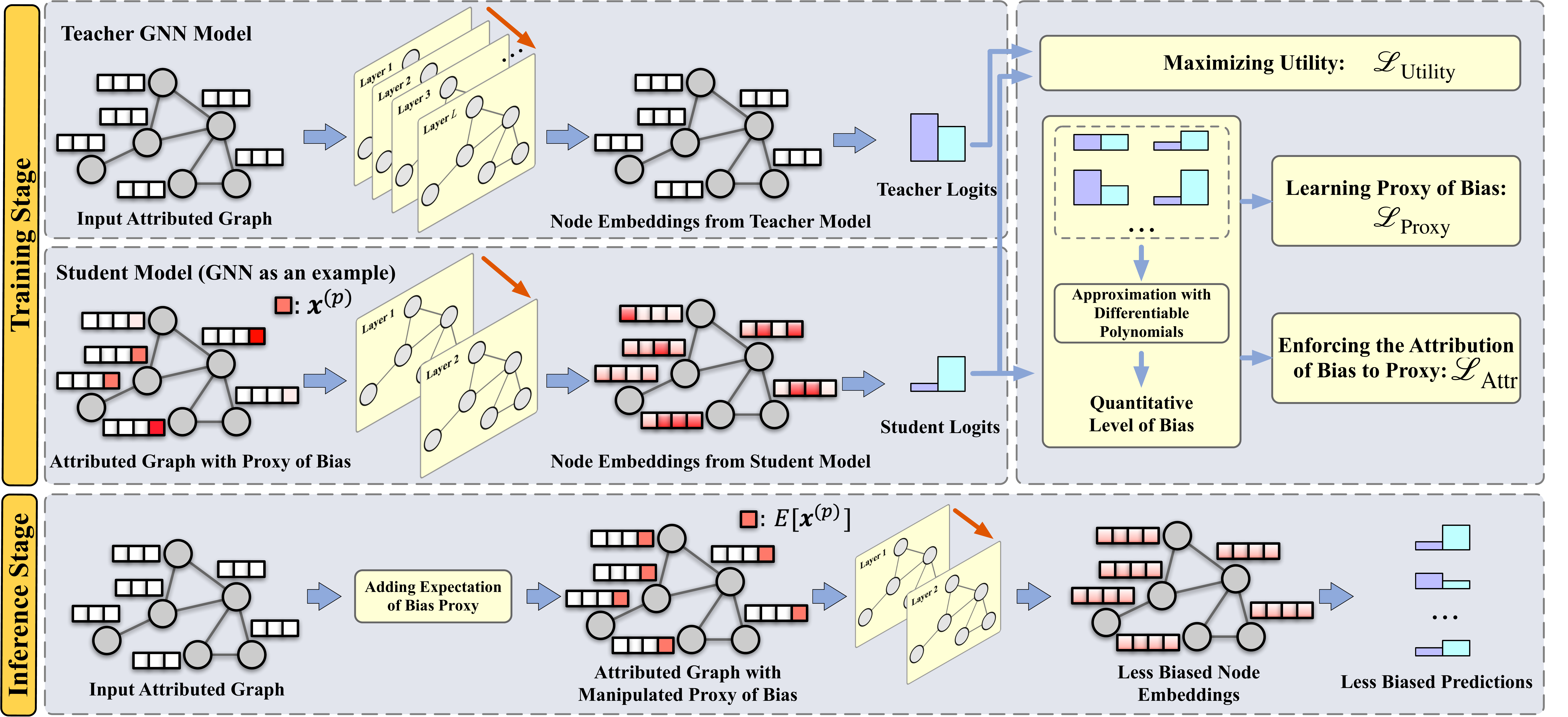}
    \vspace{-1mm}
    \caption{An overview of the proposed framework RELIANT including the training and inference stage.}
    \vspace{-5mm}
    \label{structure_overall}
\end{figure*}

\noindent \textbf{Preliminaries.}
We utilize $\mathcal{G} = \{\mathcal{V}, \mathcal{E}, \mathcal{X}\}$ to denote an attributed network (graph). Here, $\mathcal{V} = \{v_1, ..., v_n\}$ is the set of nodes, $\mathcal{E} \subseteq \mathcal{V} \times \mathcal{V}$ is the set of edges, and $\mathcal{X} = \{\mathbf{x}_1, ..., \mathbf{x}_n\}$ ($\mathbf{x}_i \in \mathbb{R}^{d}$, $1 \leq i \leq n$) is the set of node attribute vectors.
We use $\mathbf{A} \in \{0,1\}^{n \times n}$ to denote the adjacency matrix of the graph. If there is an edge from the $i$-th node to the $j$-th node, $\mathbf{A}_{i,j}=1$; otherwise $\mathbf{A}_{i,j}=0$.
Moreover, we denote the pre-trained teacher GNN model in a knowledge distillation framework as $f_{\hat{\bm{\theta}}}$ parameterized by $\hat{\bm{\theta}}$.
Here $\hat{\bm{\theta}}$ denotes the optimized $\bm{\theta}$ of the pre-trained teacher model.
Similarly, we denote the student model as $g_{\bm{\phi}}$ parameterized by $\bm{\phi}$.
We represent the optimized $\bm{\phi}$ after the training of the student model as $\hat{\bm{\phi}}$.
Without loss of generality, we consider the most widely studied node classification as the downstream task. 
For the teacher model $f_{\hat{\bm{\theta}}}(v)$, we denote the set of outcome logits, i.e., the continuous output vector corresponding to each node, as $\mathcal{\hat{Y}}^{(\text{t})} = \{\hat{\mathbf{y}}_1^{(\text{t})},\hat{\mathbf{y}}_2^{(\text{t})}, ..., \hat{\mathbf{y}}_n^{(\text{t})}\}$, where $ \hat{\mathbf{y}}_i^{(\text{t})} \in \mathbb{R}^{c}$. Here $c$ is the total number of node classes.
Correspondingly, we represent the set of outcome logits of the student model $g_{\bm{\phi}}(v)$ as $\mathcal{\hat{Y}}^{(\text{s})} = \{\hat{\mathbf{y}}_1^{(\text{s})},\hat{\mathbf{y}}_2^{(\text{s})}, ..., \hat{\mathbf{y}}_n^{(\text{s})}\}$.
For any node $v_i$, the predicted label given by the student model (denoted as $\hat{Y}^{(s)}_i$ for the $i$-th node) is determined by the largest value across all $c$ dimensions in $\hat{\mathbf{y}}_i^{(\text{s})}$.
%
% We utilize $\mathcal{Y} = \{Y_1, Y_2, ..., Y_n\}$ to specify the labels of the $n$ nodes. Correspondingly, we represent the predicted label set as $\hat{\mathcal{Y}} = \{\hat{Y}_1, \hat{Y}_2, ..., \hat{Y}_n\}$.

Based on the definitions above, we formulate the problem of \textit{Fair Knowledge Distillation for GNN-based Teacher-Student Frameworks} as follows.

% \begin{myDef1}
% \label{p1}
% \textbf{Fair Knowledge Distillation for GNN-Based Teacher-Student Frameworks.} Given an attributed network $\mathcal{G}$, a trained teacher GNN $f_{\hat{\bm{\theta}}}$, and a student model $g_{\bm{\phi}}$ to be trained, our goal is to achieve a more fair student model with similar prediction utility compared with $f_{\hat{\bm{\theta}}}$ for the node classification task.
% \end{myDef1}

\begin{myDef1}
\label{p1}
\textbf{Fair Knowledge Distillation for GNN-Based Teacher-Student Frameworks.} Given an attributed network $\mathcal{G}$ and a GNN-based teacher-student framework including a trained teacher GNN $f_{\hat{\bm{\theta}}}$ and a student model $g_{\bm{\phi}}$ to be trained, our goal is to achieve a more fair student model with similar prediction utility compared with $f_{\hat{\bm{\theta}}}$ for the node classification task.
\end{myDef1}

\section{Methodology}

In this section, we first present an overview of the proposed framework RELIANT, followed by the objective function formulation and optimization strategy.

\subsection{Workflow of RELIANT}
\label{workflow}

Here we first introduce the workflow of the proposed framework RELIANT. 
In general, we introduce the three main functionalities involved in the proposed framework RELIANT, namely \emph{maximizing the utility}, \emph{learning proxy of bias}, and \emph{enforcing the attribution of bias to the proxy}. 
We present an overview of RELIANT in Fig.~\ref{structure_overall}.
Specifically, to tackle the first challenge (gap towards fair knowledge), we propose to first learn the proxy of bias as extra input attributes for the student model to account for the exhibited bias (on training nodes), and then exclude the information of proxy during test with manipulated pseudo proxy. To tackle the second challenge (gap towards end-to-end learning), we formulate our debiasing objectives in a differentiable (w.r.t. the parameters of the student model) manner. To tackle the third challenge (gap towards generalization), we achieve debiasing in a student-agnostic manner. In other words, the proposed framework RELIANT does not rely on any specific student model structure to achieve debiasing.
% design these in a student-agnostic manner. In other words, such design does not rely on any specific student model structure.
% Finally, we discuss the strategy to achieve less biased inference with the student model.
%
We elaborate more details as follows.

% \noindent \textbf{Analysis of Exhibited Bias on Student Model.}
% In general, we consider two sources of bias that could influence the student model: (1) the input node attributes could be biased, and thus the student model could learn to make predictions based on these attributes with biased information; 
% %
% (2) the supervision information from the teacher GNN model could be biased, which leads to an optimized parameter set that make the student model tend to yield biased predictions.
% %
% A straightforward approach to tackle the first source of bias is to remove those sensitive attributes from the input network data. However, it has been proved that those non-sensitive attributes could also have dependency with the sensitive attributes. Moreover, the graph structure could also carry information about sensitive attributes. As a consequence, information about sensitive attributes cannot be naively removed from the input network data.
% %
% In this paper, we propose to learn proxy of bias as extra input attributes for the student model to account for the exhibited bias, and avoid such bias during inference via proxy manipulation.
% %

\noindent \textbf{Maximizing Utility.}
In general, existing GNN-based KD frameworks consider the GNNs with high computational costs as the teacher model, and the goal is to learn a student model with limited computational costs but similar prediction utility (e.g., accuracy in node classification tasks).
To maintain the utility of the teacher model, it is necessary to utilize the knowledge from the teacher model as the supervision signal for the training of the student.
In particular, a common approach is to utilize the output classification logits from the teacher model as the supervision signal, which we take as an example here. Specifically, we minimize the distance between the logits from the student model and the teacher model. We formally formulate the optimization goal as
\begin{align}
\min _{\bm{\phi}} \sum_{v_i \in \mathcal{V}} \gamma_{\text{d}}\left(\hat{\mathbf{y}}^{(t)}_i, \hat{\mathbf{y}}^{(s)}_i\right),
\end{align}
where $\hat{\mathbf{y}}^{(s)}_i$ and $\hat{\mathbf{y}}^{(t)}_i$ are the output logits from the student model $g_{\bm{\phi}}(v_i)$ and teacher model $f_{\hat{\bm{\theta}}}(v_i)$, respectively.
The function $\gamma_{\text{d}}(.,.)$ measures the distance between two logit vectors.
Different choices can be adopted to measure the distance, e.g., cosine distance and Euclidean distance.
Correspondingly, to maximize the prediction utility, we minimize the objective function
\begin{align}
\mathscr{L}_{\text{Utility}} (\bm{\phi}) =  \sum_{v_i \in \mathcal{V}} \gamma_{\text{d}}\left(\hat{\mathbf{y}}^{(t)}_i, \hat{\mathbf{y}}^{(s)}_i\right).
\end{align}

\noindent \textbf{Learning Proxy of Bias.}
It is worth noting that even if the sensitive attributes are removed from the input data, the student model could still exhibit bias in its predictions. The main reason is that there could exist dependencies between those sensitive attributes and non-sensitive ones. 
Moreover, the information about sensitive attributes could also be encoded in the input network structure~\cite{dong2021edits}.
As a consequence, it is difficult to prevent the student model from leveraging information about sensitive attributes.
To handle such a problem, we propose to learn the proxy of bias $\mathbf{x}_i^{(p)}$ as extra input attributes for each node $v_i$. 
Here, the rationale is that if much information about bias comes from the learned proxy instead of those encoded in the non-sensitive attributes or the network structures, then we are able to achieve less biased predictions by not using the information from such a proxy during test.
As a consequence, such a proxy of bias should account for the exhibited bias of the student model as much as possible. In other words, the exhibited bias should be largely attributed to the proxy of bias rather than the sensitive information encoded in the network data.
%
% We achieve such a goal via learning the proxy of bias in a contrastive manner.
%
More specifically, to enforce the proxy of bias contributing to the exhibited bias in the student model, we propose to maximize the exhibited bias when these proxies are taken as input into the student model together with other attributes and the network structure. We formally formulate our goal as
\begin{align}
\label{proxy_goal_1}
\max_{\mathbf{X}^{(p)}} J_{\text{Bias}}(\{ g_{\bm{\phi}}(  \gamma (v_i, \mathbf{X}^{(p)})):  i \in \mathcal{V} \}),
\end{align}
where $\gamma (.,.)$ is a function that takes a node and the proxy of bias matrix as input, and outputs the node with a concatenated node attribute vector $[\mathbf{x}_i, \mathbf{x}_i^{(p)}]$. Here $\mathbf{x}_i^{(p)}$ is the $i$-th row of $\mathbf{X}^{(p)}$.
$J_{\text{Bias}}(.)$ is a function that takes the set of logits from the student model as input and outputs a value indicating the level of exhibited bias.
%
% $\hat{\mathbf{y}}^{(s)}_i \in \hat{\mathcal{Y}}^{(s)}$ is a function of $\mathbf{x}_i^{(p)}$.
%
Nevertheless, the computation is non-differentiable under traditional fairness notions such as Statistical Parity and Equal Opportunity. Here we propose to utilize orthogonal polynomials (e.g., Legendre polynomials~\cite{dattoli2001note}) that are differentiable w.r.t. the output logits to approximate the level of bias under traditional fairness notions. This makes $J_{\text{Bias}}$ differentiable w.r.t. the learnable parameter $\bm{\phi}$.
Correspondingly, we formally give the objective function towards the goal above as
\begin{align} \label{eq:proxy}
\mathscr{L}_{\text{Proxy}}(\mathbf{X}^{(p)}) = - J_{\text{Bias}}(\{ g_{\bm{\phi}}(  \gamma (v_i, \mathbf{X}^{(p)})):  i \in \mathcal{V} \}).
\end{align}

% \noindent \textbf{Enforcing the Effectiveness of Proxy.}
\noindent \textbf{Enforcing the Attribution of Bias to the Proxy.}
Only achieving Eq.~(\ref{proxy_goal_1}) is not enough to enforce the proxy of bias largely accounting for the exhibited bias of the student model. This is because the vanilla node attributes could still contribute to the exhibited bias.
% (we consider binary classification task for simplicity) This is because the exhibited bias may come from the vanilla node attributes instead of the learned proxy.
More specifically, we denote $P(\hat{Y}^{(s)})$ as the probability of a certain classification prediction given by the student model for any specific node.
We assume that there are underlying unbiased and biased node attributes $\mathbf{X}^{(u)}$ and $\mathbf{X}^{(b)}$, respectively.
When Eq.~(\ref{proxy_goal_1}) is achieved, it is clear that $P(\hat{Y}^{(s)} | \mathbf{X}^{(u)}, \mathbf{X}^{(b)}, \mathbf{X}^{(p)})$, i.e., $P(\hat{Y}^{(s)} | \mathbf{X}, \mathbf{X}^{(p)})$, is biased. However, both $\mathbf{X}^{(b)}$ and $\mathbf{X}^{(p)}$ could be the source of the exhibited bias.
It is worth noting that our goal is to learn proxy $\mathbf{X}^{(p)}$ to account for as much of the exhibited bias as possible.
Therefore, to enforce the effectiveness of the proxy, it is necessary to ensure that the exhibited bias is attributed to the biased information from $\mathbf{X}^{(p)}$ instead of $\mathbf{X}^{(b)}$.
In other words, we need to enforce $P(\hat{Y}^{(s)} | \mathbf{X}^{(u)}, \mathbf{X}^{(b)})$ being less biased, which ensures that $\mathbf{X}^{(p)}$ accounts for the exhibited bias as much as possible.
Nevertheless, $P(\hat{Y}^{(s)} | \mathbf{X}^{(u)}, \mathbf{X}^{(b)})$ is intractable considering that the number of the input dimension number for the student model is fixed.
Hence we propose an alternative approach.
Denote the learned proxy of bias and the underlying sensitive attribute vector of any node as $\mathbf{x}^{(p)}$ and $\mathbf{s}$, respectively.
We propose to utilize a vector $\mathbb{E}[\mathbf{x}^{(p)}]$ to replace each row in $\mathbf{X}^{(p)}$ as the manipulated pseudo proxy $\tilde{\mathbf{X}}^{(p)}$. In this way, the rows in $\tilde{\mathbf{X}}^{(p)}$ are independent from $\mathbf{s}$, i.e., the information about sensitive attributes encoded in $\mathbf{X}^{(p)}$ is wiped out.
%
% If the student GNN model
%
To enforce the attribution of bias to the proxy $\mathbf{X}^{(p)}$, the predictions should be as fair as possible when the information about $\mathbf{X}^{(p)}$ is removed.
Therefore, we formulate our last optimization goal as
\begin{align}
\label{proxy_goal_2}
\min_{\bm{\phi}} J_{\text{Bias}}(\{ g_{\bm{\phi}}(  \tilde{\gamma} (v_i, \tilde{\mathbf{X}}^{(p)})):  i \in \mathcal{V} \}),
\end{align}
where $ \tilde{\gamma} (.,.)$ is a function that takes a node and the matrix $\tilde{\mathbf{X}}^{(p)}$ as input, and returns the input node with a concatenated node attribute vector $[\mathbf{x}_i, \tilde{\mathbf{x}}^{(p)}_i]$.
Here $\tilde{\mathbf{x}}^{(p)}_i$ is the $i$-th row of matrix $\tilde{\mathbf{X}}^{(p)}$.
We formally present the corresponding objective function as 
\begin{align}
    \mathscr{L}_{\text{Attr}}(\bm{\phi}) = J_{\text{Bias}}(\{ g_{\bm{\phi}}(   \tilde{\gamma} (v_i, \tilde{\mathbf{X}}^{(p)})):  i \in \mathcal{V} \}).
\end{align}

\noindent \textbf{Inference with Student Model.}
To achieve less biased inference, an ideal case is to make predictions with $P(\hat{Y}^{(s)} | \mathbf{X}^{(u)})$. However, it is difficult to explicitly extract $\mathbf{X}^{(u)}$ from $\mathbf{X}$. Instead, we argue that $P(\hat{Y}^{(s)} | \mathbf{X}^{(u)}, \mathbf{X}^{(b)}, \tilde{\mathbf{X}}^{(p)})$ exhibits similar level of bias compared with $P(\hat{Y}^{(s)} | \mathbf{X}^{(u)})$.
This is because (1) the bias exhibited by $P(\hat{Y}^{(s)} | \mathbf{X}^{(u)}, \mathbf{X}^{(b)}, \tilde{\mathbf{X}}^{(p)})$ minimally relies on $\mathbf{X}^{(b)}$ after enforcing Eq.~(\ref{proxy_goal_1}) and Eq.~(\ref{proxy_goal_2}); and (2) there is no further information about sensitive attributes encoded in $\tilde{\mathbf{X}}^{(p)}$ (as discussed above).
Consequently, we propose to utilize $g_{\bm{\phi}}(  \tilde{\gamma} (v_i, \tilde{\mathbf{X}}^{(p)}))$ to achieve less biased prediction for node $v_i$ in the inference stage.

% In such a case, we argue that $P(\hat{Y}^{(s)} | \mathbf{X}^{(u)})$ is close to $P(\hat{Y}^{(s)} | \mathbf{X}^{(u)}, \mathbf{X}^{(b)})$, and it is reasonable to utilize $P(\hat{Y}^{(s)} | \mathbf{X}, \tilde{\mathbf{X}}^{(p)})$ as an approximation of $P(\hat{Y}^{(s)} | \mathbf{X}^{(u)}, \mathbf{X}^{(b)})$.
%
% $P(\mathbf{y}^{(s)} | \mathbf{X}^{(u)}) \approx P(\mathbf{y}^{(s)} | \mathbf{X}^{(u)}, \mathbf{X}^{(b)}) \approx  P(\mathbf{y}^{(s)} | \mathbf{X}, \mathbb{E}(\mathbf{X}^{(p)}))$.
%
% Correspondingly, we propose to utilize $g_{\bm{\phi}}(  \tilde{\gamma} (v_i, \tilde{\mathbf{X}}^{(p)}))$ to make the prediction corresponding to node $v_i$ in the inference stage. 

\begin{table*}[]
\vspace{-3mm}
\caption{The basic information about the real-world datasets adopted for experimental evaluation. Sens. denotes the semantic meaning of sensitive attribute.}
\vspace{-2mm}
\label{datasets}
% \footnotesize
\small
\centering
\setlength{\extrarowheight}{-0.9pt}
\begin{tabular}{lcccccc}
\hline
\textbf{Dataset}           & \textbf{Recidivism}  & \textbf{Credit Defaulter}     & \textbf{DBLP}  & \textbf{DBLP-L}           \\
\hline
\textbf{\# Nodes}          & 18,876     & 30,000        & 39,424         & 129,726                                      \\
\textbf{\# Edges}         & 321,308   & 1,436,858           & 52,460          & 591,039                             \\
\textbf{\# Attributes}     & 18     & 13           & 5,693          & 5,693                                          \\
\textbf{Avg. degree}      & 34.0  & 95.8         & 1.3           & 4.6                                       \\
\textbf{Sens.}           & Race   & Age   & Continent of Affiliation   & Continent of Affiliation        \\
\textbf{Label}         & Bail Decision   & Future Default   & Research Field  & Research Field      \\
\hline
\end{tabular}
\vspace{-0.6em}
\end{table*}

% \subsection{Objective Function Formulation}
% In this subsection, we introduce the objective function formulation of the proposed RELIANT.

% \noindent \textbf{Objective of Utility.}

% $\mathscr{L}_1$

% \noindent \textbf{Objective of Learning Bias Proxy.}

% $\mathscr{L}_2$

% \noindent \textbf{Objective of Debiasing.}

% $\mathscr{L}_3$

\section{Optimization Objectives \& Strategy}

We present the optimization objectives of RELIANT followed by the training strategy in this section.

\noindent \textbf{Optimization Objectives.} Based on our discussions above, here we present a summary of the optimization objectives for the proposed RELIANT. 
First, to optimize the parameter $\bm{\phi}$, we formally formulate a unified objective function as
\begin{align}
\label{phi_loss}
\mathscr{L}_{\bm{\phi}} = \mathscr{L}_{\text{Utility}} (\bm{\phi})  + \lambda \cdot \mathscr{L}_{\text{Attr}}(\bm{\phi}).
\end{align}
Here $\lambda$ serves as a hyper-parameter controlling the effect of debiasing the student model.
Second, to optimize the learnable proxy of bias $\mathbf{X}^{(p)}$, we formally present the objective function as
\begin{align}
\label{xp_loss}
\mathscr{L}_{\mathbf{X}^{(\text{P})}} = \mathscr{L}_{\text{Proxy}}(\mathbf{X}^{(p)}).
\end{align}

\noindent \textbf{Optimization Strategy.}
To train the proposed framework RELIANT, we propose to optimize the parameter $\bm{\phi}$ and learnable proxy of bias $\mathbf{X}^{(p)}$ in an alternating manner. We present the algorithmic routine of RELIANT in Algorithm~\ref{algorithm1}.

\vspace{-2mm}
\begin{algorithm}[H]
  \footnotesize
    \caption{Fair Knowledge Distillation for GNNs}
    \begin{algorithmic}[1]
    \label{algorithm1}
      \REQUIRE 
$\mathcal{G}$: the graph data; 
$f_{\hat{\bm{\theta}}}$: the trained teacher GNN model; 
$g_{\bm{\phi}}$: the student model; 
\ENSURE 
$g_{\hat{\bm{\phi}}}$: the optimized student model; $\mathbf{X}^{(p)}$: the proxy of bias matrix; \\
\STATE Randomly initialize $\mathbf{X}^{(p)}$; 
\WHILE{stop training condition not satisfied}
\STATE Compute $\mathscr{L}_{\bm{\phi}}$ according to Eq.~(\ref{phi_loss}); 
\STATE Update $\bm{\phi}$ with $\frac{\partial \mathscr{L}_{\bm{\phi}}}{\partial \bm{\phi}}$; 
\STATE Compute $\mathscr{L}_{\mathbf{X}^{(p)}}$ according to Eq.~(\ref{xp_loss}); 
\STATE Update $\mathbf{X}^{(p)}$ with $\frac{\partial \mathscr{L}_{\mathbf{X}^{(p)}}}{\partial \mathbf{X}^{(p)}}$; 
\ENDWHILE
\RETURN $g_{\hat{\bm{\phi}}}$ and $\mathbf{X}^{(p)}$;
\end{algorithmic}
\end{algorithm}
\vspace{-1.6em}

\section{Experimental Evaluations}

\label{experiments}

In this section, we will first introduce the downstream learning task and adopted real-world datasets, followed by the backbone models, baseline methods, and evaluation metrics. Next, we present the implementation details of the models. Finally, we discuss the evaluation results of the proposed RELIANT.
In particular, we aim to answer the following research questions through experiments: \textbf{RQ1:} How well can RELIANT balance the utility and fairness of the student model compared with other baselines? \textbf{RQ2:} To what extent each component of RELIANT contributes to the overall debiasing performance? \textbf{RQ3:} How will the choice of the hyper-parameter $\lambda$ affect the performance of RELIANT?

\begin{table*}[t!]
% \vspace{-10mm}
\caption{The experimental results based on node classification accuracy and $\Delta_{\text{SP}}$. We use "(T)" and "(S)" suffixes to represent the teacher model and the student model, respectively. Here Vanilla(S) denotes the student model trained with the vanilla KD framework; One-Hot(S) represents the student model trained with the one-hot bias proxy; RELIANT(S) is the student model trained with our proposed model. $\uparrow$ denotes the larger, the better; while $\downarrow$ denotes the opposite. All quantitative results are presented in percentages. The best results are in \textbf{Bold}.}
\vspace{-2mm}
\label{sp_results}
\centering
\small
\setlength{\extrarowheight}{-0.9pt}
\setlength\tabcolsep{7.5pt}
\begin{tabular}{lclcccc}
\hline

                                                                     &                                       &            & \textbf{DBLP} & \textbf{DBLP-L} & \textbf{Credit} & \textbf{Recidivism} \\
                                                                     \hline
\multirow{8}{*}{\begin{tabular}[c]{@{}l@{}}\textbf{CPF}\\ \textbf{+GCN}\end{tabular}}  & \multirow{4}{*}{\textbf{Accuracy ($\uparrow$)}}                  & \textbf{GCN(T)}    & 92.37 $\pm$ 0.06 & 94.20 $\pm$ 0.09 & 76.39 $\pm$ 0.48 & \textbf{93.68 $\pm$ 0.21} \\
                                                                     &                                       & \textbf{Vanilla(S)} & \textbf{93.14 $\pm$ 0.10}     & \textbf{94.30 $\pm$ 0.04}       & \textbf{77.85 $\pm$ 0.10}       & 89.41 $\pm$ 0.12           \\
                                                                     &                                       & \textbf{One-Hot(S)}  & 93.04 $\pm$ 0.34     & 94.16 $\pm$  0.02       & 77.65 $\pm$ 0.10       & 89.15 $\pm$  0.37 \\
                                                                     &                                       & \textbf{RELIANT(S)} & 92.70 $\pm$ 0.40     & 94.07 $\pm$  0.18       & 77.82  $\pm$ 0.45       & 88.88  $\pm$ 0.57 \\
                                                                     \cline{2-7}
                                                                     & \multirow{4}{*}{\textbf{$\Delta_{\text{SP}}$ ($\downarrow$)}} & \textbf{GCN(T)} & 7.66 $\pm$ 0.26    & 7.33 $\pm$0.44  & 15.81 $\pm$0.40   & 6.10 $\pm$0.05 \\
                                                                     &                                       & \textbf{Vanilla(S)} & 8.55 $\pm$ 0.50  & 7.16 $\pm$ 0.16       & 14.90 $\pm$ 0.89       & 6.85 $\pm$ 0.05 \\
                                                                     &                                       & \textbf{One-Hot(S)}  & 7.97 $\pm$ 0.63     & 7.46 $\pm$ 0.24       & 13.80 $\pm$ 0.32       & 6.78 $\pm$ 0.51 \\
                                                                     &                                       & \textbf{RELIANT(S)} & \textbf{2.27 $\pm$ 1.00}     & \textbf{3.09 $\pm$ 0.36}       & \textbf{10.28 $\pm$ 1.86}       & \textbf{4.06 $\pm$ 0.64} \\
                                                                     \hline
\multirow{8}{*}{\begin{tabular}[c]{@{}l@{}}\textbf{CPF}\\ \textbf{+SAGE}\end{tabular}} & \multirow{4}{*}{\textbf{Accuracy ($\uparrow$)}}                  & \textbf{SAGE(T)}    & 92.57 $\pm$ 0.28 & 94.10 $\pm$ 0.25 & 77.88 $\pm$ 0.06 & \textbf{89.71 $\pm$ 0.14} \\
                                                                     &                                       & \textbf{Vanilla(S)} & \textbf{93.25 $\pm$ 0.15}     & \textbf{94.97 $\pm$ 0.10}       & 77.97 $\pm$ 0.26       & 89.20 $\pm$ 0.11 \\
                                                                     &                                       & \textbf{One-Hot(S)}  & 93.07 $\pm$ 0.10     & 94.32 $\pm$ 0.07       & 78.01 $\pm$ 0.23       & 89.11 $\pm$ 0.29  \\
                                                                     &                                       & \textbf{RELIANT(S)} & 92.91 $\pm$ 0.51     & 94.17 $\pm$ 0.93       & \textbf{78.28 $\pm$ 0.36}       & 88.85 $\pm$ 0.27 \\
                                                                     \cline{2-7}
                                                                     & \multirow{4}{*}{\textbf{$\Delta_{\text{SP}}$ ($\downarrow$)}} & \textbf{SAGE(T)}    & 8.32 $\pm$0.24  & 7.81 $\pm$0.08  & 14.08 $\pm$ 1.37   & 6.50 $\pm$0.39       \\
                                                                     &                                       & \textbf{Vanilla(S)} & 8.29 $\pm$ 0.85     & 7.02 $\pm$ 0.13       & 13.44 $\pm$ 5.23       & 4.41 $\pm$ 0.43  \\
                                                                     &                                       & \textbf{One-Hot(S)}  & 8.01 $\pm$ 0.25     & 7.52 $\pm$ 0.32       & 16.86 $\pm$ 3.86       & 6.62 $\pm$ 0.38 \\
                                                                     &                                       & \textbf{RELIANT(S)} & \textbf{2.01 $\pm$ 1.21}     & \textbf{2.97 $\pm$ 0.61}       & \textbf{10.06 $\pm$ 1.70}       & \textbf{3.94 $\pm$ 0.60} \\
                                                                     \hline
\multirow{8}{*}{\begin{tabular}[c]{@{}l@{}}\textbf{AKD}\\ \textbf{+GCN}\end{tabular}}  & \multirow{4}{*}{\textbf{Accuracy ($\uparrow$)}}                  & \textbf{GCN(T)}    & \textbf{92.37 $\pm$ 0.06} & \textbf{94.20 $\pm$ 0.09} & \textbf{76.39 $\pm$ 0.48} & \textbf{93.68 $\pm$ 0.21} \\
                                                                     &                                       & \textbf{Vanilla(S)} & 92.06 $\pm$ 0.16     & 94.07 $\pm$ 0.11       & 76.35 $\pm$ 0.31       & 92.08 $\pm$ 0.29 \\
                                                                     &                                       & \textbf{One-Hot(S)}  & 91.55 $\pm$ 0.40     & 94.07 $\pm$ 0.04       & 75.65 $\pm$ 0.75  & 92.07 $\pm$ 0.03 \\
                                                                     &                                       & \textbf{RELIANT(S)} & 91.39 $\pm$ 0.24     & 93.98 $\pm$ 0.08       & 75.64 $\pm$ 0.06       & 91.21 $\pm$ 0.14 \\
                                                                     \cline{2-7}
                                                                     & \multirow{4}{*}{\textbf{$\Delta_{\text{SP}}$ ($\downarrow$)}} & \textbf{GCN(T)}    & 7.66 $\pm$0.26 & 7.33 $\pm$0.44  & 15.81$\pm$0.40   & 6.10 $\pm$0.05 \\
                                                                     &                                       & \textbf{Vanilla(S)} & 7.87 $\pm$ 0.25     & 6.79 $\pm$ 0.10       & 13.61 $\pm$ 2.00       & 6.54 $\pm$ 0.17  \\
                                                                     &                                       & \textbf{One-Hot(S)}  & 7.39 $\pm$ 0.35     & 6.72 $\pm$ 0.19       & 14.30 $\pm$ 0.24       & 6.44 $\pm$ 0.32  \\
                                                                     &                                       & \textbf{RELIANT(S)} & \textbf{3.66 $\pm$ 1.09}     & \textbf{5.18 $\pm$ 0.16}       & \textbf{8.47 $\pm$ 1.92}       & \textbf{5.70 $\pm$ 0.18} \\
                                                                     \hline
\multirow{8}{*}{\begin{tabular}[c]{@{}l@{}}\textbf{AKD}\\ \textbf{+SAGE}\end{tabular}} & \multirow{4}{*}{\textbf{Accuracy ($\uparrow$)}}                  & \textbf{SAGE(T)}    & \textbf{92.57 $\pm$ 0.28} & 94.10 $\pm$ 0.25 & 77.88 $\pm$ 0.06 & \textbf{89.71 $\pm$ 0.14} \\
                                                                     &                                       & \textbf{Vanilla(S)} & 92.23 $\pm$ 0.07     & 94.45 $\pm$ 0.03       & 78.10 $\pm$ 0.24       & 89.67 $\pm$ 0.07 \\
                                                                     &                                       & \textbf{One-Hot(S)}  & 92.31 $\pm$ 0.06     & \textbf{94.52 $\pm$ 0.11}       & 78.24 $\pm$ 0.45       & 89.60 $\pm$ 0.12 \\
                                                                     &                                       & \textbf{RELIANT(S)} & 92.07 $\pm$ 0.07     & 94.28 $\pm$ 0.06       & \textbf{78.60 $\pm$ 0.33}       & 88.87 $\pm$ 0.31  \\
                                                                     \cline{2-7}
                                                                     & \multirow{4}{*}{\textbf{$\Delta_{\text{SP}}$ ($\downarrow$)}} & \textbf{SAGE(T)}    & 8.32 $\pm$0.24 & 7.81 $\pm$0.08  & 14.08 $\pm$ 1.37   & 6.50 $\pm$0.39 \\
                                                                     &                                       & \textbf{Vanilla(S)} & 7.53 $\pm$ 0.29     & 7.34 $\pm$ 0.41       & 14.41 $\pm$ 0.15       & 6.24 $\pm$ 0.20 \\
                                                                     &                                       & \textbf{One-Hot(S)}  & 7.72 $\pm$ 0.44     & 7.26 $\pm$ 0.36       & 11.69 $\pm$ 0.93       & 6.18 $\pm$ 0.30 \\
                                                                     &                                       & \textbf{RELIANT(S)} & \textbf{4.91 $\pm$ 0.64}      & \textbf{4.05 $\pm$ 0.14}       & \textbf{5.00 $\pm$ 1.63}       & \textbf{6.06 $\pm$ 0.26} \\
                                                                     \hline
\end{tabular}
\vspace{-1.5em}
\end{table*}

\subsection{Experimental Settings}

Here we introduce the settings for our experimental evaluation.

\noindent \textbf{Downstream Task \& Real-world Datasets.}
We adopt the widely studied node classification as the downstream task in this paper. We adopt four real-world datasets for the experimental evaluations, including two widely used network datasets (Recidivism~\cite{jordan2015effect,agarwal2021towards} and Credit Defaulter~\cite{yeh2009comparisons,agarwal2021towards}) and two newly constructed ones based on real-world data (DBLP and DBLP-L).
%
% namely Recidivism~\cite{jordan2015effect}, Credit Defaulter~\cite{yeh2009comparisons}, DBLP, and DBLP-L.
%
In Recidivism, nodes are defendants released on bail, and edges denote the connections between defendants computed from their past criminal records. Here the sensitive feature is race, and we aim to classify if a certain defendant is unlikely to commit a crime after bail.
In Credit Defaulter, nodes are credit card users, and edges are the connections between these users. Here we consider the age period of these users as their sensitive feature, and we aim to predict the future default of credit card payments.
Additionally, we also construct two co-author networks, namely DBLP and DBLP-L based on AMiner network~\cite{tang2008arnetminer}, which is a co-author network collected from computer science bibliography.
Specifically, we first filter out the nodes in AMiner network with incomplete information.
Then we adopt two different approaches to sample a connected network from the filtered dataset: DBLP is a subgraph sampled with random walk, while DBLP-L is the largest connected component of the filtered AMiner network.
In both datasets, nodes represent the researchers in different fields, and edges denote the co-authorship between researchers. The sensitive attribute is the continent of the affiliation each researcher belongs to, and we aim to predict the primary research field of each researcher.
The detailed statistics of these four datasets are in Table~\ref{datasets}.

\noindent \textbf{KD Framework Backbones \& Teacher GNNs.}
To evaluate the capability of RELIANT in generalizing to different GNN-based KD backbones, here we adopt two representative KD frameworks designed for compressing GNNs, namely CPF~\cite{yang2021extract} and GraphAKD~\cite{he2022compressing}. In general, CPF minimizes the distribution distance between the logits from teacher and student to provide supervision information for the student, while GraphAKD utilizes adversarial training to achieve knowledge distillation for the student. The student model of CPF and GraphAKD is PLP~\cite{yang2021extract} and SGC~\cite{wu2019simplifying}, respectively.
For each KD framework, we adopt two types of GNNs (including GCN~\cite{DBLP:conf/iclr/KipfW17} and GraphSAGE~\cite{hamilton2017inductive}) as the teacher GNN.

\noindent \textbf{Baselines.}
To the best of our knowledge, this is the first study on how to mitigate the bias exhibited in GNN-based KD frameworks. In experiments, we adopt the student model yielded by the vanilla KD framework as our first baseline. For our second baseline, we replace the learnable proxy of bias with a naive proxy for the input of the KD framework. Specifically, we utilize one-hot vectors as the naive proxy for different demographic subgroups during training, where the one-hot vector flags the membership of different nodes. We replace all proxy vectors during inference with an averaged proxy vector across all instances.
Here, the rationale is that more distinguishable attributes are easier for deep learning models to learn during training, and these one-hot vectors serve as an "easier" indicator of biased information. In this way, if these one-hot proxy accounts for the exhibited bias of the student model after training, then the exhibited bias could also be mitigated during inference, where such information is wiped out.

\noindent \textbf{Evaluation Metrics.}
We evaluate the performance of the compressed GNN models (i.e., the output student model of KD frameworks) from two perspectives, namely utility and fairness. Specifically, in terms of utility, we adopt the node classification accuracy as the corresponding metric; in terms of fairness, we adopt two traditional metrics $\Delta_{\text{SP}}$ and $\Delta_{\text{EO}}$.
Here $\Delta_{\text{SP}}$ measures the bias level (of predictions) under the fairness notion of Statistical Parity, while $\Delta_{\text{EO}}$ measures the bias level under the notion of Equal Opportunity.
%
% We present the quantitative results of $\Delta_{\text{EO}}$ in Appendix~\ref{supp_eo} due to space limit.
%
See online version of this paper for other results in Appendix due to space limit.

\noindent \textbf{Implementation Details.}
RELIANT is implemented in PyTorch~\cite{paszke2017automatic} and optimized with Adam optimizer~\cite{kingma2014adam}. In our experiments, the learning rate 
is chosen in $\{10^{-2},10^{-3},10^{-4}\}$ and the training epoch number is set as 1,000 for CPF and 600 for GraphAKD. Experiments are carried out on an Nvidia RTX A6000, and the reported numerical results are averaged across three different runs. We introduce more details in Appendix.

\subsection{Effectiveness of RELIANT}

\label{experiments_balance}

Here we aim to answer \textbf{RQ1}. Specifically, we evaluate our proposed framework RELIANT on two KD backbones, namely CPF and GraphAKD. For each KD backbone, we adopt two different GNNs (GCN and GraphSAGE) to evaluate the capability of our proposed framework in generalizing to different GNNs. We compare the corresponding performances between the teacher GNN model and the student models trained with three different frameworks (i.e., the vanilla KD framework, the KD framework with the one-hot proxy of bias, and our proposed RELIANT).
We present quantitative results on node classification accuracy and $\Delta_{\text{SP}}$ in Table~\ref{sp_results}. In addition, we also perform experiments based on Equal Opportunity (see Appendix), where we have consistent observations.
We make the following observations from Table~\ref{sp_results}.
\begin{itemize}[topsep=1pt]
\setlength{\itemsep}{2pt}
\setlength{\parsep}{2pt}
\setlength {\parskip}{2pt}
    \item From the perspective of prediction utility, student models trained with all three KD frameworks achieve comparable performances with the teacher model. This implies that effective knowledge distillation can be achieved by all three KD frameworks. 
    \item From the perspective of bias mitigation, the student models trained with the vanilla KD frameworks inherit and even exaggerate the exhibited bias from the teacher GNN model in all cases. Training the student models with the one-hot proxy can mitigate bias in most cases. Compared with the student models trained with the vanilla KD framework and the one-hot proxy, RELIANT consistently exhibits less bias w.r.t. Statistical Parity.
    \item Based on the performance of RELIANT in both perspectives, RELIANT achieves effective debiasing for the student model but still maintains comparable model utility with the teacher model. Therefore, we argue that RELIANT achieves a satisfying balance between debiasing and maintaining utility.
\end{itemize}

\subsection{Ablation Study}
\begin{figure}[!t]
\centering
% \vspace{-4mm}
        \begin{subfigure}[t]{0.24\textwidth}
        \small
        \includegraphics[width=0.98\textwidth]{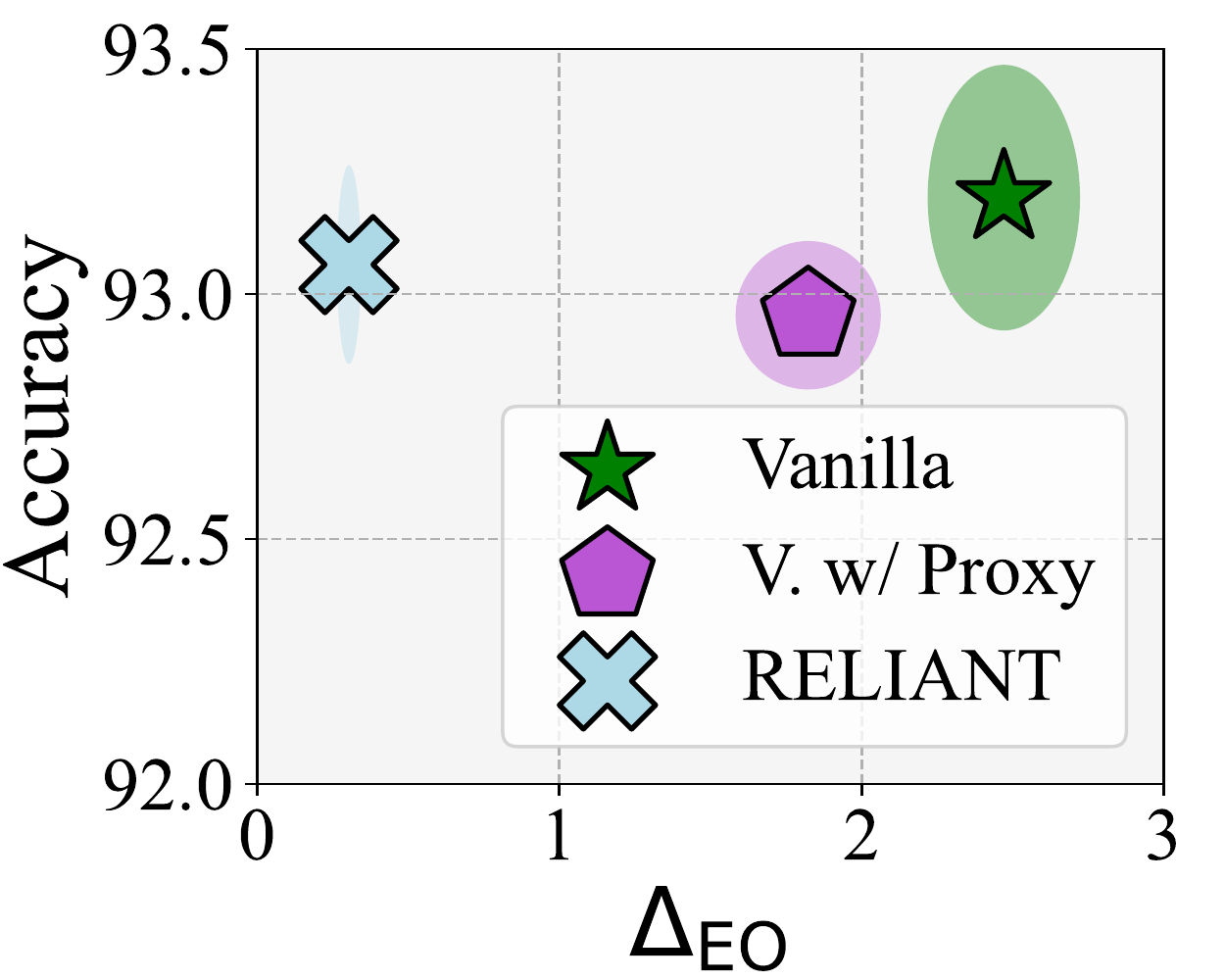}
        % \vspace{-1mm}
            \caption[Network2]%
            % {{\footnotesize $\Delta G_{\text{SP}}$ on Income}} 
            {{\footnotesize GraphAKD on Credit.}} 
            \label{ablation1}
        \end{subfigure}
        \hspace{-1mm}
        \begin{subfigure}[t]{0.24\textwidth}
        \small
        \includegraphics[width=0.98\textwidth]{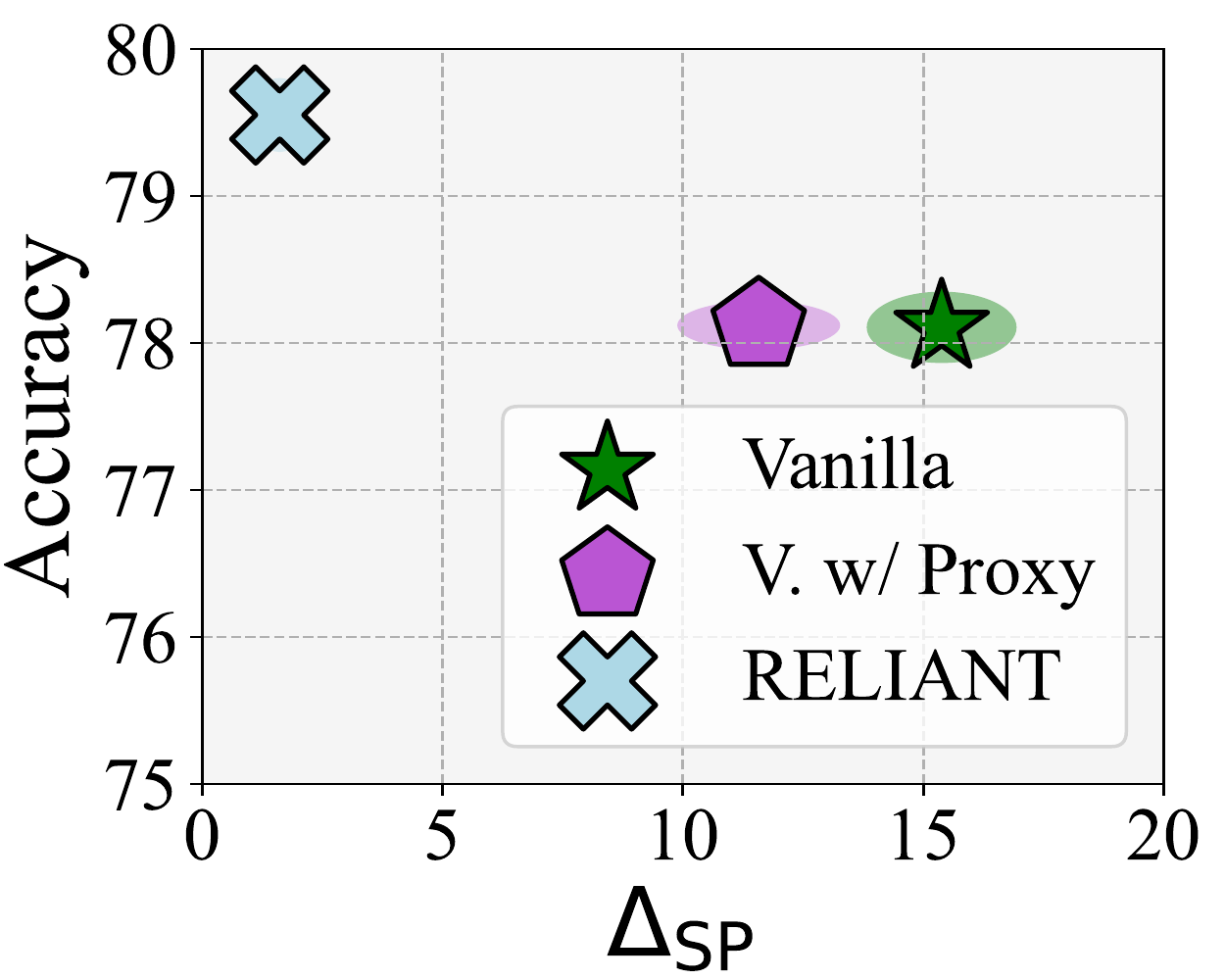}
        % \vspace{-1mm}
            \caption[Network2]%
            % {{\footnotesize $\Delta G_{\text{SP}}$ on Income}} 
            {{\footnotesize CPF on DBLP.}} 
            \label{ablation2}
        \end{subfigure}
         \vspace{-2mm}
    \caption{Ablation study of RELIANT. "Vanilla" denotes the student model trained with the original KD framework, while "V. w/ Proxy" represents the student model trained under the KD framework with only learning the proxy of bias.}
    % \vspace{-1mm}
    \label{ablation_study}
\end{figure}

We aim to answer \textbf{RQ2} in this subsection. Specifically, for each framework, we evaluate to what extent the two modules of RELIANT (including \textit{learning proxy of bias} and \textit{enforcing the attribution of bias to the proxy}) contribute to the performance of the student model.
We present the results in Fig.~\ref{ablation_study}.
Here, Fig.~\ref{ablation1} is the performance of accuracy vs. $\Delta_{\text{SP}}$ from CPF based on the DBLP-L dataset, while Fig.~\ref{ablation2} is the performance of accuracy vs. $\Delta_{\text{EO}}$ from GraphAKD based on the Recidivism dataset. 
Notably, we also have similar observations under other settings. We make the following observations.
\begin{itemize}[topsep=1pt]
\setlength{\itemsep}{2pt}
\setlength{\parsep}{2pt}
\setlength {\parskip}{2pt}
    \item From the perspective of prediction utility, we observe that the prediction utility is comparable among all three cases. This corroborates that both modules exert limited influence on the node classification accuracy.
    \item From the perspective of bias mitigation, adding the module of \textit{learning proxy of bias} to the vanilla KD framework brings limited bias mitigation. This is because the bias could also come from the non-sensitive node attributes (as discussed in Section~\ref{workflow}). After the module of \textit{enforcing the attribution of bias to the proxy} is added together with \textit{learning proxy of bias}, RELIANT is then able to achieve satisfying performance on bias mitigation. 
\end{itemize}

\begin{figure}[!t]
\centering
\vspace{-5mm}
        \begin{subfigure}[t]{0.241\textwidth}
        \small
        \includegraphics[width=0.98\textwidth]{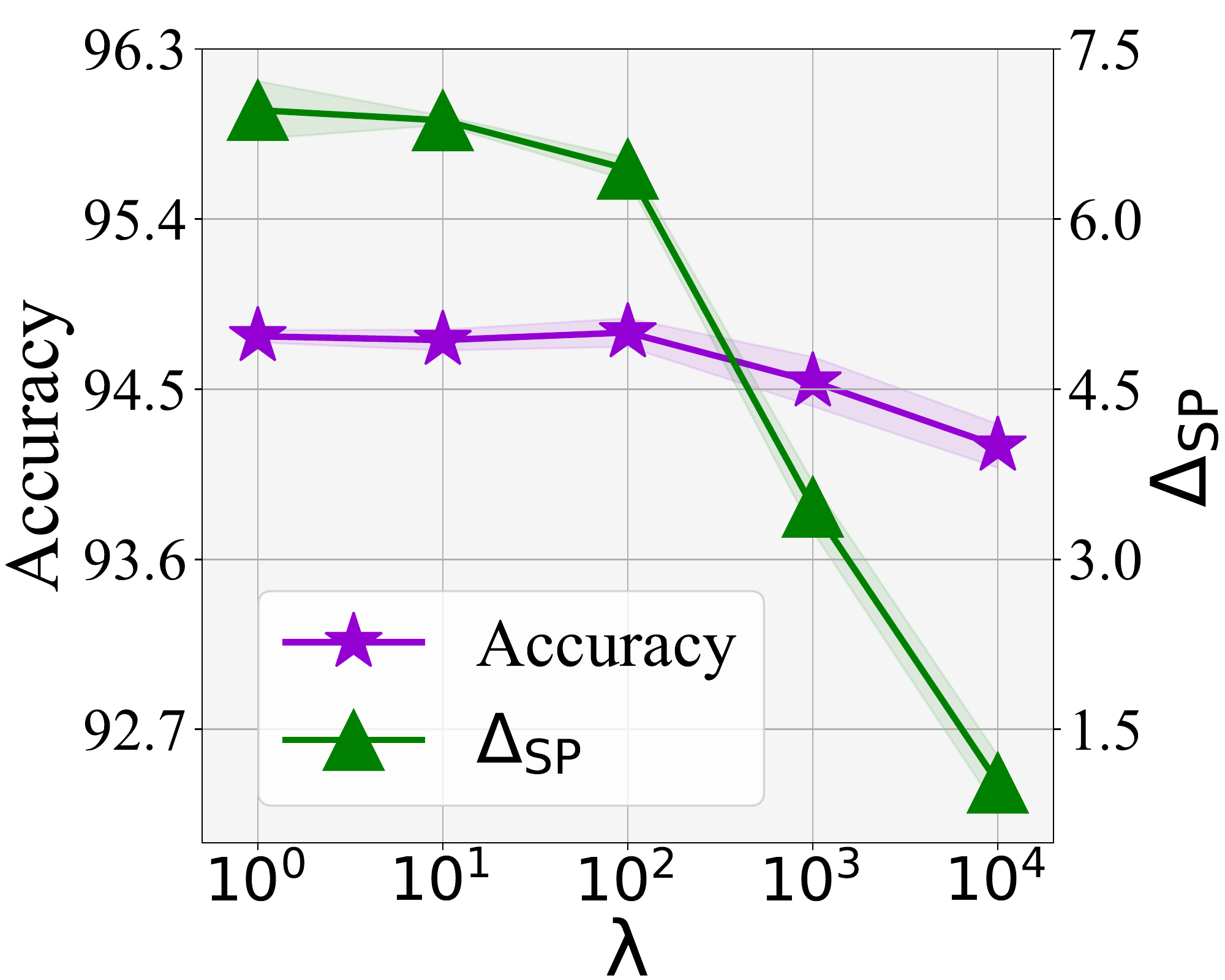}
        % \vspace{-1mm}
            \caption[Network2]%
            % {{\footnotesize $\Delta G_{\text{SP}}$ on Income}} 
            {{\footnotesize CPF on DBLP-L.}} 
            \label{parameter_study1}
        \end{subfigure}
        % \hspace{2mm}
                        \begin{subfigure}[t]{0.241\textwidth}
        \small
        \includegraphics[width=0.98\textwidth]{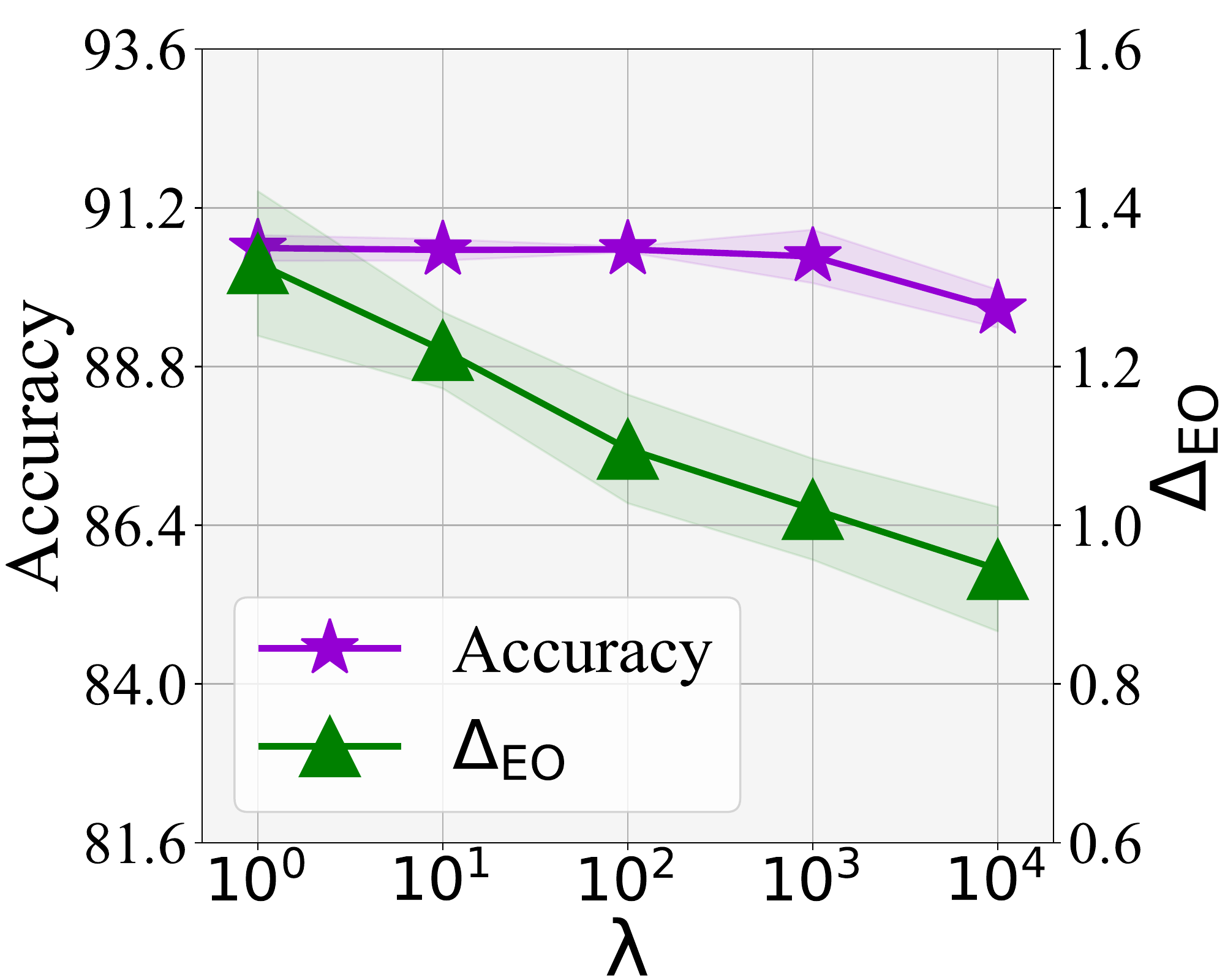}
        % \vspace{-1mm}
            \caption[Network2]%
            {{\footnotesize GraphAKD on Recidivism.}} 
            \label{parameter_study2}
        \end{subfigure} 
        \vspace{-2mm}
    \caption{Parameter sensitivity of $\lambda$ based on two different KD backbones on two real-world datasets. We also have similar observations on other datasets.}
    \vspace{-8mm}
    \label{parameter_study}
\end{figure}

\subsection{Parameter Sensitivity}
We answer \textbf{RQ3} by studying the tendency of model utility and exhibited bias w.r.t. the change of hyper-parameter $\lambda$. Here $\lambda$ controls the effect of $\mathscr{L}_{\text{Attr}}$. More specifically, we vary $\lambda$ in \{$10^0$, $10^1$, $10^2$, $10^3$, $10^4$\}, and we present the corresponding tendency of node classification accuracy and the exhibited bias of the trained student model with RELIANT in Fig.~\ref{parameter_study}. Here, Fig.~\ref{parameter_study1} is based on 
the Credit dataset under GraphAKD, while Fig.~\ref{parameter_study2} is based on the DBLP dataset under CPF. We also have similar observations on other datasets.
We make the following observations from Fig.~\ref{parameter_study}.
\begin{itemize}[topsep=1pt]
\setlength{\itemsep}{2pt}
\setlength{\parsep}{2pt}
\setlength {\parskip}{2pt}
    \item From the perspective of prediction utility, the node classification accuracies on both datasets and KD backbones do not exhibit apparent reduction when the value of $\lambda$ increases from $1$ to $10^4$. This verifies that the prediction utility is not sensitive to $\lambda$.
    \item From the perspective of bias mitigation, the student model exhibits less bias when $\lambda$ increases from $1$ to $10^4$. Specifically, when $\lambda$ is relatively small (e.g., $1$), the learned proxy of bias only partially accounts for the exhibited bias; when the value of $\lambda$ increases, more bias is then attributed to the learned proxy. Considering the balance between model utility and bias mitigation, a recommended range of $\lambda$ is between $10^2$ and $10^3$.
\end{itemize}

\section{Related Works}

\noindent \textbf{Algorithmic Fairness in GNNs.}
Most existing works promoting the algorithmic fairness of GNNs focus either on \textit{Group Fairness}~\cite{dwork2012fairness} or \textit{Individual Fairness}~\cite{zemel2013learning}.
Specifically, group fairness is defined based on a set of pre-defined sensitive attributes (e.g., gender and race). These sensitive attributes divide the whole population into different demographic subgroups. Group fairness requires that each subgroup should receive their fair share of interest according to the output GNN predictions~\cite{mehrabi2021survey}.
Various explorations have been made towards achieving a higher level of group fairness for GNNs~\cite{dong2022fairness}. 
Decoupling the output predictions from sensitive attributes via adversarial learning is one of the most popular approaches among existing works~\cite{fairview,dai2021say}.
Other common strategies include reformulating the objective function with fairness regularization~\cite{fan2021fair,navarin2020learning}, rebalancing the number of intra-group edges between two demographic subgroups~\cite{dong2021edits,li2020dyadic}, deleting nodes or edges that contribute the most to the exhibited bias~\cite{dong2023bind,dong2022structural}, etc.
On the other hand, individual fairness does not rely on any sensitive attributes. Instead, individual fairness requires that similar nodes (in the input space) should be treated similarly (in the output space)~\cite{dwork2012fairness}.
%
% Such a criterion is usually formulated with Lipschitz condition~\cite{dwork2012fairness}.
%
To fulfill individual fairness in GNNs, adding fairness-aware regularization terms to the optimization objective is the most widely adopted approach~\cite{dong2021individual,song2022guide}.
%
% However, current explorations of individual fairness still remain limited.

\noindent \textbf{Knowledge Distillation.}
In recent years, knowledge distillation has been proven to be effective in compressing the model but still maintaining similar model prediction performance~\cite{gou2021knowledge}. Correspondingly, it has been widely adopted in a plethora of applications, including visual recognition~\cite{wu2020learning}, natural language processing~\cite{haidar2019textkd,JiaoYSJCL0L20}, etc. The main idea of knowledge distillation is to transfer the knowledge of a computationally expensive teacher model to a light student model, and thus the student model is able to fit in platforms with limited computing resources~\cite{he2022compressing,joshi2021representation}.
It is worth noting that such a strategy is also proved to be effective in compressing GNNs~\cite{yang2021extract,he2022compressing,joshi2021representation}. Consequently, there is growing research attention on utilizing knowledge distillation to compress GNNs for more efficient inference.
For example, encouraging the student model to yield similar output to the teacher GNN via regularization is proved to be effective~\cite{he2022compressing}. In addition, adversarial learning is also a popular technique to obtain light-weighted but accurate student models~\cite{he2022compressing}.
However, most of these frameworks for GNNs do not have fairness consideration. Hence the student model tends to be influenced by biased knowledge from the teacher GNN.
Different from existing works, we develop a generalizable knowledge distillation framework that explicitly considers fairness in GNNs but still maintains the utility of GNN predictions.

\section{Conclusion}

Despite the success of Knowledge Distillation (KD) in compressing GNNs, most existing works do not consider fairness. Hence the student model trained with the KD framework tends to inherit and even exaggerate the bias from the teacher GNN. In this paper, we take initial steps towards learning less biased student models for GNN-based KD frameworks. Specifically, we first formulate a novel problem of fair knowledge distillation for GNN-based teacher-student frameworks, then propose a framework named RELIANT to achieve a less biased student model. Notably, the design of RELIANT is agnostic to the specific structures of teacher and student models. Therefore, it can be easily adapted to different KD approaches for debiasing. Extensive experiments demonstrate the effectiveness of RELIANT in fulfilling fairness for GNN compression with KD.

\section{Acknowledgments}
\label{ack}
This work is supported by the National Science Foundation under grants IIS-2006844, IIS-2144209, IIS-2223768, IIS-2223769, CNS-2154962, CMMI-2125326, BCS-2228533, and BCS-2228534, the JP Morgan Chase Faculty Research Award, and the Cisco Faculty Research Award. We would like to thank the anonymous reviewers for their constructive feedback.

\clearpage

\linespread{0.95}
\bibliographystyle{acm}  % acm
\bibliography{ref_renewed}

% \begin{thebibliography}{99}
% \end{thebibliography}

\clearpage
\appendix

\linespread{0.98}

\section{Reproducibility}
\label{repro}

In this section, we introduce the reproducibility of the presented experiments as a supplement of Section~\ref{experiments}. More specifically, we first introduce the experimental settings in detail, followed by the implementation details of our proposed framework RELIANT, GNNs, KD backbones, and the baseline debiasing framework. We finally introduce several key packages and their corresponding versions used in our implementations.
The code will be released upon acceptance.

\subsection{Experimental Settings.}
The implementation of our experiments could mainly be divided into three parts, well-trained teacher GNNs, knowledge distillation backbones, and our RELIANT module. Generally, our experiments are implemented on an Nvidia RTX A6000. We write our experiment code in PyTorch \cite{paszke2019pytorch} framework, and use Adam \cite{kingma2014adam} optimizer to learn the model parameters. We run all experiments three times and record the average and the standard deviation where the random seeds are chosen in \{0,10,100,1000,10000\}.

\subsection{Implementation of RELIANT.}
We implement RELIANT in PyTorch and use Adam as the optimizer of the learnable proxy. For results shown in \Cref{sp_results} and \Cref{eo_results}, we set the proxy learning rate as $10^{-2}$ and weight decay as $10^{-2}$ for CPF and set the proxy learning rate as $10^{-2}$ and weight decay as $5\times 10^{-4}$ for GraphAKD. For all the results, we search for the optimal coefficient $\lambda$ in the set \{1,10,100,1000,10000\}.

\subsection{Implementation of Graph Neural Networks.}
For the training of the teacher GNN models, we use the code in the CPF framework where the details of implementation setting are shown in \Cref{tab:teacher_setting}.

\subsection{Implementation of KD Frameworks.}\label{KD_implementation}
%
% mark out the open source code link with \footnote{https:xxx}
In our experimental setting, the teacher model is fixed during the training stage. Hence we first train a teacher GNN with the parameters shown in \Cref{tab:teacher_setting}. After obtaining a well-trained teacher GNN, we use it as supervision to train the student model. Details of student training are shown as follows.
\begin{itemize}[topsep=1pt]
\setlength{\itemsep}{2pt}
\setlength{\parsep}{2pt}
\setlength {\parskip}{2pt}
    \item For CPF, we follow the implementation in \cite{yang2021extract}, where we use PLP as the student model. For the training settings of the student model, we set the maximum epoch as 1,000, early stopping as 500, layer number as 5, feature dropout as 0.8, edge weight dropout as 0.2, learning rate in $\{10^{-2},10^{-3},10^{-4}\}$, and weight decay as $10^{-2}$.
    \item For GraphAKD, we follow the implementation in \cite{he2022compressing}, where we utilize SGC~\cite{wu2019simplifying} as the student model. For detailed training settings, we set the maximum epoch as 600, layer number as 3, learning rate in $\{10^{-2},10^{-3},10^{-4}\}$, weight decay as $5\times 10^{-4}$, and dropout as 0.5.
\end{itemize}

\subsection{Implementation of the Baseline Debiasing Framework.}
For the vanilla KD frameworks, we follow the settings in \Cref{KD_implementation} to directly implement the KD framework.
For the baselines with the one-hot proxy for bias, we use the one-hot vector of sensitive attributes as the constant proxy. It is added to the input data before training and is fixed during the training stage. Hence there is no loss term (as Eq. \ref{eq:proxy} shows) for the constant proxy. The training settings of the one-hot baseline are the same as Vanilla since the constant proxy vectors do not contain any extra parameters.

% \begin{itemize}
% \setlength{\itemsep}{2pt}
% \setlength{\parsep}{2pt}
% \setlength {\parskip}{2pt}
%     \item \textbf{Vanilla}. For Vanilla baseline, we follow the settings in \Cref{KD_implementation} to directly implement the KD framework.
%     \item \textbf{One-Hot}. For One-Hot baseline, we use the one-hot vector of sensitive attributes as the constant proxy. It is added to the input data before training and is fixed during the training stage. Hence there is no loss term (as \Cref{eq:proxy} shows) for the constant proxy. The training settings of One-Hot baseline are the same as Vanilla because constant proxy does not contain any extra parameters.
% \end{itemize}

\begin{table}[t]
    % \vspace{-3mm}
    \caption{Experimental settings of teacher GNN models.}\label{tab:teacher_setting}
    \vspace{-2mm}
    \centering
    \begin{tabular}{lcc}
    \toprule
    Hyperparameter & GCN & GraphSAGE \\
    \midrule
    Layer & 3 & 3 \\
    Hidden Dimension & 64 & 128 \\
    Epoch & 1000 & 1000 \\
    Early Stopping & 500 & 500 \\
    Learning Rate & $10^{-2}$ & $10^{-2}$ \\
    Weight Decay & $10^{-3}$ & $5\times 10^{-4}$ \\
    Dropout & 0.8 & 0.5 \\
    % Aggregation Type & N.A. & GCN \\
    \bottomrule
    \end{tabular}
    \label{tab:my_label}
\end{table}

\subsection{Packages Required for Implementations.}
% \noindent \textbf{Packages Required for Implementations.}
%
We list the key packages and corresponding versions in our implementations as below. 
\begin{itemize}[topsep=1pt]
\setlength{\itemsep}{0pt}
\setlength{\parsep}{0pt}
\setlength {\parskip}{0pt}
    \item Python == 3.7.10
    \item torch == 1.8.1
    \item torch-cluster == 1.5.9
    \item torch-geometric == 1.4.1
    % \item torch-scatter == 2.0.6
    \item torch-sparse == 0.6.9
    % \item cuda == 11.0
    \item numpy == 1.20.0
    % \item tensorboard == 1.13.1
    \item networkx == 2.5.1
    \item scikit-learn == 0.24.1
    \item pandas==1.2.3
    \item scipy==1.4.1
\end{itemize}

\begin{figure}[!t]
\centering
    \vspace{-3mm}
        \begin{subfigure}[t]{0.49\textwidth}
        \small
        \includegraphics[width=0.98\textwidth]{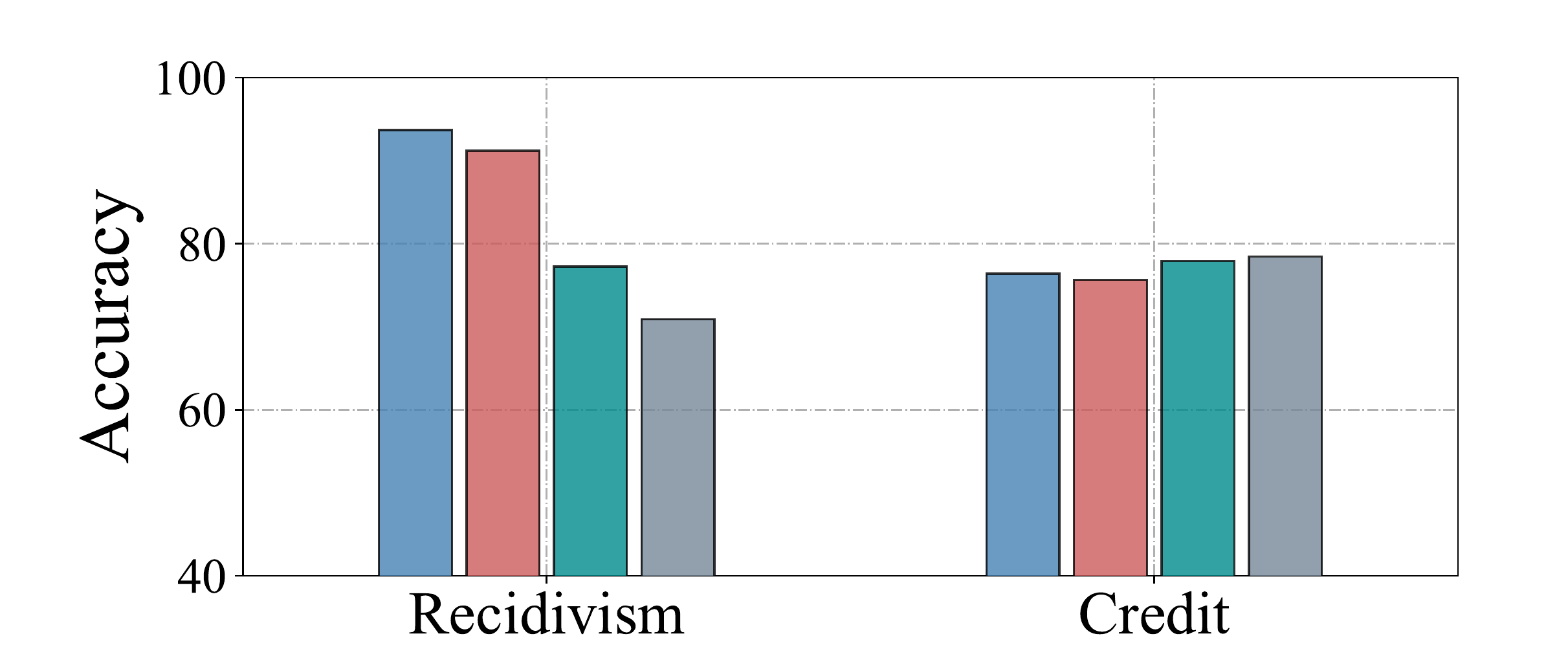}
        % \vspace{-1mm}
            \caption[Network2]%
            % {{\footnotesize $\Delta G_{\text{SP}}$ on Income}} 
            {{\footnotesize Comparison on the node classification accuracy.}} 
            \vspace{-1mm}
            \label{sup_comp1}
        \end{subfigure}
                        \begin{subfigure}[t]{0.49\textwidth}
        \small
        \includegraphics[width=0.98\textwidth]{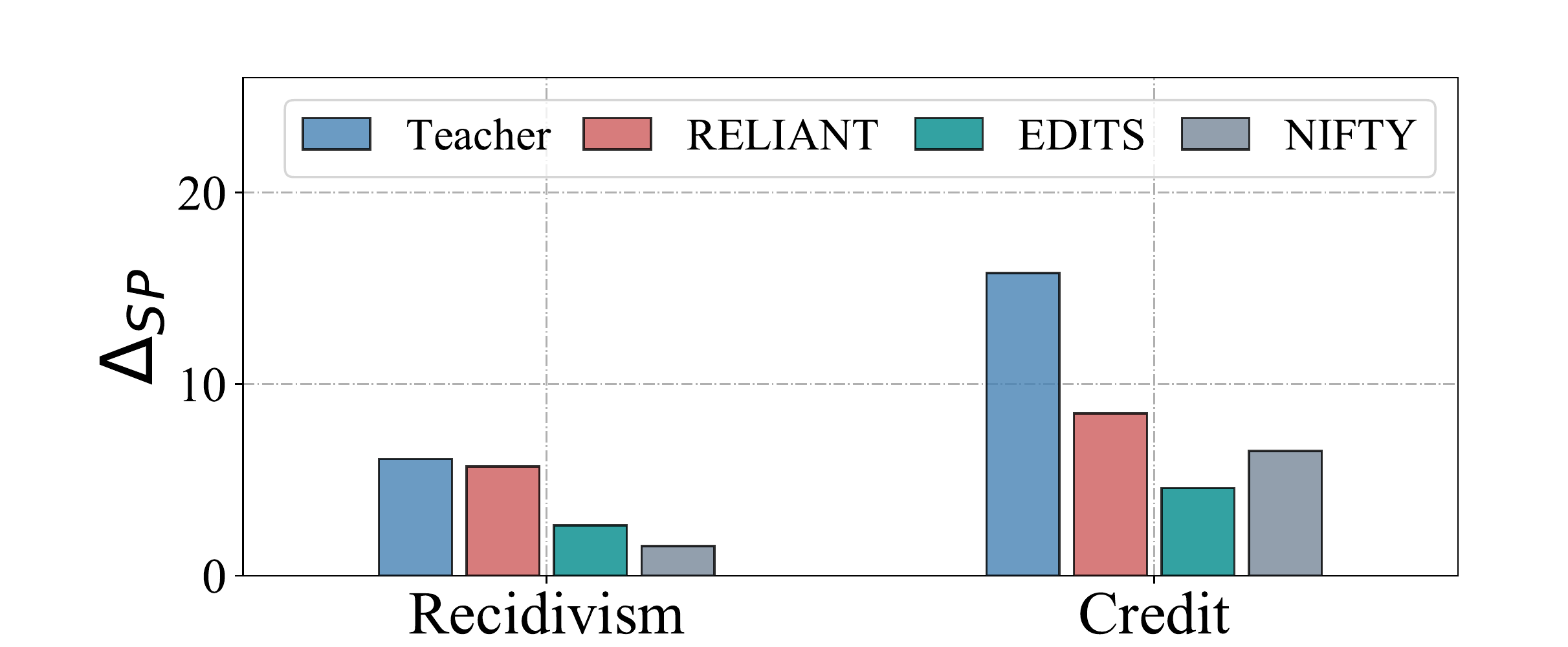}
        % \vspace{-1mm}
            \caption[Network2]%
            {{\footnotesize Comparison on $\Delta_{\text{SP}}$.}}    
            \label{sup_comp2}
        \end{subfigure} 
        \vspace{-3mm}
    \caption{Performance comparison on balancing prediction utility and bias mitigation between the proposed framework RELIANT and other state-of-the-art methods that directly debias GNNs. All numerical results are in percentage.}
    \vspace{-5mm}
    \label{sup_comp}
\end{figure}

\section{Complementary Experiments}

In this section, we perform experiments as a supplement of Section~\ref{experiments}. Specifically, we first present the performance of RELIANT over prediction utility and fairness under another widely studied fairness notion -- Equal Opportunity. Then, we perform experiments to compare the performance of RELIANT on balancing the GNN prediction utility and fairness with state-of-the-art methods that directly debias GNNs.

\subsection{Effectiveness of RELIANT}

\label{supp_eo}

Here we present complementary experiments to answer \textbf{RQ1}, where the fairness notion is instantiated with Equal Opportunity (measured with $\Delta_{\text{EO}}$). 
Here we adopt the same settings as those in Section~\ref{experiments_balance}.
We compare the performances between the teacher GNN model and the student models trained with three different framework variants, including the vanilla KD framework, the KD framework with the one-hot proxy of bias, and our proposed RELIANT.
We present quantitative results on node classification accuracy and $\Delta_{\text{EO}}$ in Table~\ref{eo_results}.
We make the following observations from Table~\ref{eo_results}.
\begin{itemize}[topsep=1pt]
\setlength{\itemsep}{2pt}
\setlength{\parsep}{2pt}
\setlength {\parskip}{2pt}
    \item From the perspective of prediction utility, all student models trained with the adopted three KD frameworks are able to achieve comparable performances with the teacher model. This corroborates that all three KD frameworks are capable of achieving effective knowledge distillation.
    \item From the perspective of bias mitigation, the student models trained with the vanilla KD frameworks inherit or exaggerate the bias from the teacher GNN in all cases. Compared with the student models trained with the vanilla KD framework and the one-hot proxy, RELIANT consistently exhibits less bias under the fairness notion of Equal Opportunity. The student model trained with RELIANT even exhibits less bias than the teacher model in certain cases, which further corroborates the effectiveness of RELIANT in training less biased student models.
    \item According to the performance of RELIANT in both perspectives above, RELIANT is proved to achieve effective debiasing for the student model but maintains comparable prediction utility compared with the teacher model. Therefore, we argue that RELIANT achieves a satisfying balance between debiasing and maintaining the prediction utility.
\end{itemize}

\subsection{RELIANT vs. GNN-Debiasing Methods}

\label{comp_debias_gnn}

In this subsection, we perform experiments and compare the performance of RELIANT with other state-of-the-art GNN debiasing methods on balancing the prediction utility and bias mitigation.

\noindent \textbf{Baselines.}
Here we choose two state-of-the-art GNN debiasing methods as our baselines, namely EDITS~\cite{dong2021edits} and NIFTY~\cite{agarwal2021towards}.
EDITS is a recent GNN debiasing method that learns less biased network data in the pre-processing stage. After debiasing, the network data will be fed into the GNN model for evaluation. NIFTY is another recent GNN-based debiasing framework that achieves bias mitigation in the processing stage. During training, node representations are learned to be invariant to the sensitive attributes after counterfactual perturbations.

\noindent \textbf{Backbones.}
Here we choose the most widely used GCN as the backbone GNN model for all methods. We choose GraphAKD as the KD backbone of the proposed RELIANT. It is also worth noting that we also have similar observations with other GNN backbones. 

\noindent \textbf{Discussion.} We present the performance comparison results on node classification accuracy and $\Delta_{\text{SP}}$ in Fig.~\ref{sup_comp}.
We make the following observations.
\begin{itemize}[topsep=1pt]
\setlength{\itemsep}{2pt}
\setlength{\parsep}{2pt}
\setlength {\parskip}{2pt}
    \item From the perspective of prediction utility, RELIANT keeps comparable to the teacher GCN model, while other debiasing methods bear different levels of prediction utility corruption. Therefore, RELIANT achieves satisfying performance in maintaining the prediction utility among all methods.
    \item From the perspective of bias mitigation, RELIANT is able to achieve comparable debiasing performance with other baselines when all models bear similar prediction utility (e.g., on Credit dataset); when baselines outperform RELIANT on bias mitigation, there is also much more prediction utility sacrifice (e.g., on Recidivism dataset). Considering that debiasing the student model with biased supervision is much more difficult than directly debiasing GNNs, we argue that the performances of RELIANT in both cases should be considered satisfying.
    \item According to the performance of RELIANT in both perspectives above, we argue that RELIANT  achieves comparable performance with other state-of-the-art GNN debiasing approaches, which further corroborates its satisfying performance on balancing the prediction utility and bias mitigation.
\end{itemize}

\begin{table*}[]
\vspace{-10mm}
\caption{The experimental results based on node classification accuracy and $\Delta_{\text{EO}}$. We use "(T)" and "(S)" suffixes to represent the teacher model and the student model, respectively. Here Vanilla(S) denotes the student model trained with the vanilla KD framework; One-Hot(S) represents the student model trained with the one-hot bias proxy; RELIANT(S) is the student model trained with our proposed model. $\uparrow$ denotes the larger, the better; while $\downarrow$ denotes the opposite. All quantitative results are presented in percentages. The best results are in \textbf{Bold}.}
\label{eo_results}
\centering
% \small
\setlength{\extrarowheight}{2.0pt}
\setlength\tabcolsep{5pt}
\begin{tabular}{lclcccc}
\hline

                                                                     &                                       &            & \textbf{DBLP} & \textbf{DBLP-L} & \textbf{Credit} & \textbf{Recidivism} \\
                                                                     \hline
\multirow{8}{*}{\begin{tabular}[c]{@{}l@{}}\textbf{CPF}\\ \textbf{+GCN}\end{tabular}}  & \multirow{4}{*}{\textbf{Accuracy ($\uparrow$)}}                  & \textbf{GCN(T)}    & 92.37 $\pm$ 0.06 & \textbf{94.20 $\pm$ 0.09} & 76.39 $\pm$ 0.48 & \textbf{93.68 $\pm$ 0.21} \\
                                                                     &                                       & \textbf{Vanilla(S)} & \textbf{93.24 $\pm$ 0.18}     & 94.15 $\pm$ 0.04       & 77.58 $\pm$ 0.20       & 89.36 $\pm$ 0.16  \\
                                                                     &                                       & \textbf{One-Hot(S)})  & 93.07 $\pm$ 0.35     & 94.15 $\pm$ 0.07       & \textbf{77.65 $\pm$ 0.10}       & 89.38 $\pm$ 0.15  \\
                                                                     &                                       & \textbf{RELIANT(S)} & 93.20 $\pm$ 0.12     & 94.15 $\pm$ 0.08       & 77.00 $\pm$ 1.57       & 89.30 $\pm$ 0.19 \\
                                                                     \cline{2-7}
                                                                     & \multirow{4}{*}{\textbf{$\Delta_{\text{EO}}$ ($\downarrow$)}} & \textbf{GCN(T)}    & 2.31 $\pm$  0.13     & 2.29 $\pm$ 0.34      & 12.63 $\pm$ 0.24       & 0.52 $\pm$ 0.06          \\
                                                                     &                                       & \textbf{Vanilla(S)} & 2.56 $\pm$ 0.11     & 2.34 $\pm$ 0.29       & 11.56 $\pm$ 0.38       & 1.25 $\pm$ 0.19 \\
                                                                     &                                       & \textbf{One-Hot(S)}  & 2.21 $\pm$ 0.32     & 2.17 $\pm$ 0.26       & 10.69 $\pm$ 0.31       & 0.96 $\pm$ 0.28 \\
                                                                     &                                       & \textbf{RELIANT(S)} & \textbf{0.42 $\pm$ 0.21}     & \textbf{1.08 $\pm$ 0.10}       & \textbf{6.02 $\pm$ 4.78}       & \textbf{0.35 $\pm$ 0.12} \\
                                                                     \hline
\multirow{8}{*}{\begin{tabular}[c]{@{}l@{}}\textbf{CPF}\\ \textbf{+SAGE}\end{tabular}} & \multirow{4}{*}{\textbf{Accuracy ($\uparrow$)}}                  & \textbf{SAGE(T)}    & 92.57 $\pm$ 0.28 & 94.10 $\pm$ 0.25 & 77.88 $\pm$ 0.06 & \textbf{89.71 $\pm$ 0.14} \\
                                                                     &                                       & \textbf{Vanilla(S)} & 93.22 $\pm$ 0.03     & \textbf{94.37 $\pm$ 0.08}       & \textbf{78.30 $\pm$ 0.23}       & 89.15 $\pm$ 0.27 \\
                                                                     &                                       & \textbf{One-Hot(S)}  & 93.13 $\pm$ 0.11     & 94.36 $\pm$ 0.06       & 78.01 $\pm$ 0.23       & 88.98 $\pm$ 0.55 \\
                                                                     &                                       & \textbf{RELIANT(S)} & \textbf{93.24 $\pm$ 0.09}     & 94.32 $\pm$ 0.06       & 78.11 $\pm$ 0.40       & 89.01 $\pm$ 0.26 \\
                                                                     \cline{2-7}
                                                                     & \multirow{4}{*}{\textbf{$\Delta_{\text{EO}}$ ($\downarrow$)}} & \textbf{SAGE(T)}    & 2.51 $\pm$ 0.33     & 2.67 $\pm$ 0.19      & 11.05 $\pm$ 0.71       & 0.86 $\pm$ 0.03   \\
                                                                     &                                       & \textbf{Vanilla(S)} & 2.83 $\pm$ 0.34     & 2.00 $\pm$ 0.18       & 11.07 $\pm$ 4.61       & 1.17 $\pm$ 0.11 \\
                                                                     &                                       & \textbf{One-Hot(S)}  & 2.16 $\pm$ 0.27     & 2.05 $\pm$ 0.21       & 12.73 $\pm$ 2.29       & 1.23 $\pm$ 0.08 \\
                                                                     &                                       & \textbf{RELIANT(S)} & \textbf{0.63 $\pm$ 0.42}     & \textbf{0.86 $\pm$ 0.18}       & \textbf{6.72 $\pm$ 4.49}       & \textbf{0.51 $\pm$ 0.25} \\
                                                                     \hline
\multirow{8}{*}{\begin{tabular}[c]{@{}l@{}}\textbf{AKD}\\ \textbf{+GCN}\end{tabular}}  & \multirow{4}{*}{\textbf{Accuracy ($\uparrow$)}}                  & \textbf{GCN(T)}    & \textbf{92.37 $\pm$ 0.06} & \textbf{94.20 $\pm$ 0.09} & 76.39 $\pm$ 0.48 & \textbf{93.68 $\pm$ 0.21} \\
                                                                     &                                       & \textbf{Vanilla(S)} & 92.12 $\pm$ 0.09     & 94.06 $\pm$ 0.06       & \textbf{78.12 $\pm$ 0.65}       & 92.29 $\pm$ 0.06  \\
                                                                     &                                       & \textbf{One-Hot(S)}  & 91.68 $\pm$ 0.28     & 93.98 $\pm$ 0.13       & 77.87 $\pm$ 0.48       & 92.28 $\pm$ 0.13 \\
                                                                     &                                       & \textbf{RELIANT(S)} & 91.69 $\pm$ 0.19     & 94.09 $\pm$ 0.12       & 77.88 $\pm$ 0.82       & 92.46 $\pm$ 0.09 \\
                                                                     \cline{2-7}
                                                                     & \multirow{4}{*}{\textbf{$\Delta_{\text{EO}}$ ($\downarrow$)}} & \textbf{GCN(T)}    & 2.31 $\pm$ 0.13     & 2.29 $\pm$ 0.34      & 12.63 $\pm$ 0.24       & \textbf{0.52 $\pm$ 0.06}   \\
                                                                     &                                       & \textbf{Vanilla(S)} & 2.76 $\pm$ 0.33     & 1.88 $\pm$ 0.08       & 8.26 $\pm$ 3.41       & 0.82 $\pm$ 0.17 \\
                                                                     &                                       & \textbf{One-Hot(S)}  & 2.69 $\pm$ 0.28     & 1.87 $\pm$ 0.17       & 8.43 $\pm$ 5.08       & 0.97 $\pm$ 0.45 \\
                                                                     &                                       & \textbf{RELIANT(S)} & \textbf{1.79 $\pm$ 0.31}     & \textbf{1.43 $\pm$ 0.09}       & \textbf{4.96 $\pm$ 3.77}       & 0.66 $\pm$ 0.21 \\
                                                                     \hline
\multirow{8}{*}{\begin{tabular}[c]{@{}l@{}}\textbf{AKD}\\ \textbf{+SAGE}\end{tabular}} & \multirow{4}{*}{\textbf{Accuracy ($\uparrow$)}}                  & \textbf{SAGE(T)}  &  \textbf{92.57 $\pm$ 0.28} & 94.10 $\pm$ 0.25 & 77.88 $\pm$ 0.06 & 89.71 $\pm$ 0.14 \\
                                                                     &                                       & \textbf{Vanilla(S)} & 92.23 $\pm$ 0.07     & 94.45 $\pm$ 0.03       & 78.10 $\pm$ 0.24       & 90.56 $\pm$ 0.14 \\
                                                                     &                                       & \textbf{One-Hot(S)}  & 92.31 $\pm$ 0.06     & \textbf{94.52 $\pm$ 0.11}       & 78.24 $\pm$ 0.45       & \textbf{90.85 $\pm$ 0.20} \\
                                                                     &                                       & \textbf{RELIANT(S)} & 92.15 $\pm$ 0.16     & 94.42 $\pm$ 0.05       & \textbf{79.08 $\pm$ 0.29}       & 90.00 $\pm$ 0.64 \\
                                                                     \cline{2-7}
                                                                     & \multirow{4}{*}{\textbf{$\Delta_{\text{EO}}$ ($\downarrow$)}} & \textbf{SAGE(T)}    & 2.51 $\pm$ 0.33     & 2.67 $\pm$ 0.19      & 11.05 $\pm$ 0.71       & \textbf{0.86 $\pm$ 0.03}     \\
                                                                     &                                       & \textbf{Vanilla(S)} & 2.06 $\pm$ 0.06     & 2.23 $\pm$ 0.23       & 10.56 $\pm$ 0.43       & 1.61 $\pm$ 0.39 \\
                                                                     &                                       & \textbf{One-Hot(S)}  & 2.21 $\pm$ 0.39     & 2.11 $\pm$ 0.21       & 8.38 $\pm$ 0.73       & 1.10 $\pm$ 0.37 \\
                                                                     &                                       & \textbf{RELIANT(S)} & \textbf{1.60 $\pm$ 0.45}     & \textbf{1.89 $\pm$ 0.21}       & \textbf{2.33 $\pm$ 0.80}       & 0.91 $\pm$ 0.22 \\
                                                                     \hline
\end{tabular}
\end{table*}

\end{document}